\documentclass[runningheads]{llncs}

\usepackage{eccv}

\usepackage{eccvabbrv}

\usepackage{graphicx}
\usepackage{booktabs}
\usepackage{makecell}
\usepackage{arydshln} 
\usepackage{bbm}
\usepackage{subcaption} 
\usepackage{float}
\usepackage{caption}  % <- 중요!
\usepackage{afterpage}
\usepackage{algorithm}
\usepackage{algorithmic}
\usepackage{colortbl}
\usepackage{xcolor}
\usepackage{pifont}
\usepackage{multirow}
\usepackage{amsmath} % align 환경 등
\usepackage{array}   % array 환경
\usepackage{enumitem}
\usepackage{xspace}
\usepackage{siunitx}

\usepackage[accsupp]{axessibility}  % Improves PDF readability for those with disabilities.

\usepackage[breaklinks,colorlinks,citecolor=eccvblue]{hyperref}
\usepackage{orcidlink}
\newcommand{\method}{Anti-Prompt\xspace} % or 

\begin{document}
% ---------------------------------------------------------------
% TODO REVIEW: Replace with your title
\title{\method: Image Protection against Text-Guided Image-to-Video Generation} 

% TODO REVIEW: If the paper title is too long for the running head, you can set
% an abbreviated paper title here. If not, comment out.
\titlerunning{Anti-Prompt}

% TODO FINAL: Replace with your author list. 
% Include the authors' OCRID for the camera-ready version, if at all possible.
\author{Yeonghwan Song\inst{1} \quad
Chanhui Lee\inst{2} \quad
Jinsoo Park\inst{2} \quad
Jeany Son\inst{2}\thanks{Corresponding author.}
}
% TODO FINAL: Replace with an abbreviated list of authors.
\authorrunning{Song \etal}
% First names are abbreviated in the running head.
% If there are more than two authors, 'et al.' is used.

% TODO FINAL: Replace with your institution list.
% \institute{$^1$AIGS, GIST, South Korea \quad \quad $^2$GSAI, POSTECH, South Korea\\
% \institute{AI Graduate School, Gwangju Institute of Science and Technology, South Korea \and Graduate School of AI, Pohang University of Science and Technology, South Korea\\
% \institute{Gwangju Institute of Science and Technology (GIST), South Korea \and Pohang University of Science and Technology (POSTECH), South Korea\\a
\institute{$^1$ GIST \quad \quad $^2$ POSTECH\\
\url{https://yeonghwansong.github.io/Anti-prompt/}}

\makeatletter
\makeatother

\maketitle
\vspace{-0.5cm}
\begin{figure}
\centering
\includegraphics[width=1\linewidth]{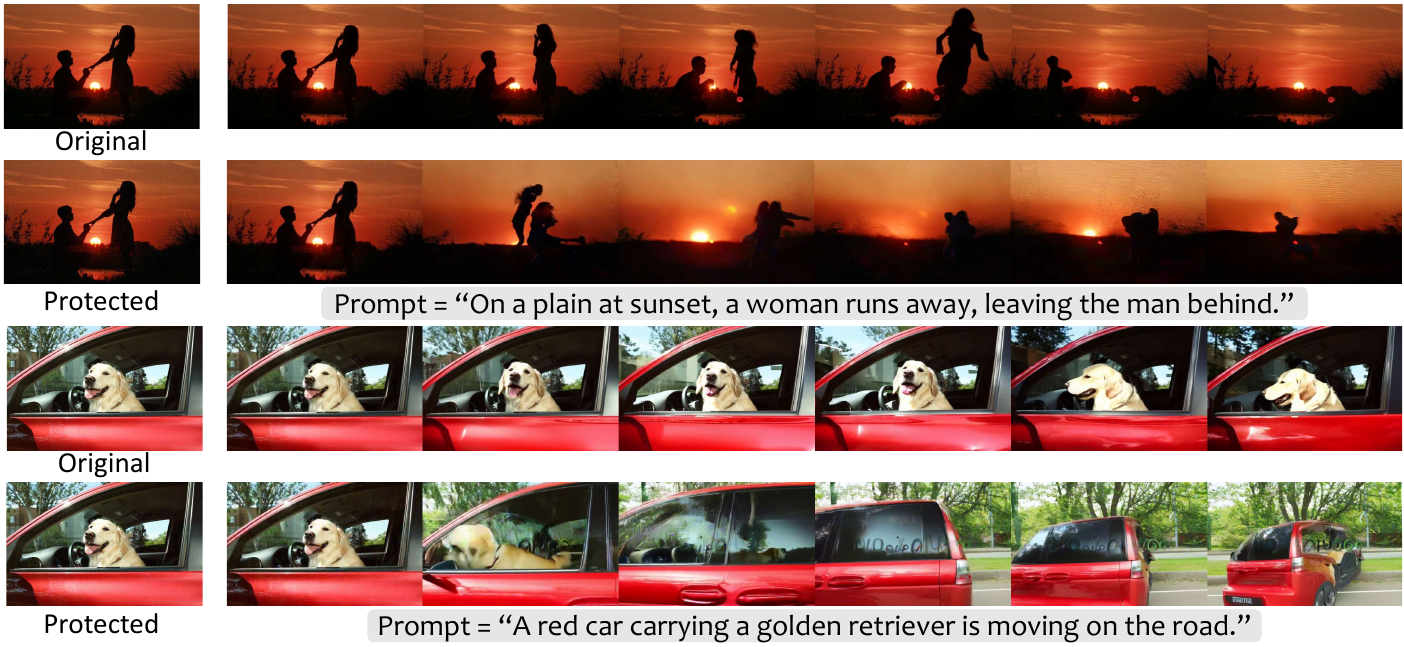}
\caption{\textbf{Text-guided I2V generation and protection.}
CogVideoX~\cite{cog} generates videos from the original image (top) and our protected image (bottom).}
\vspace{-1cm}
\label{fig:teaser}
\end{figure}

\begin{abstract}
Recent advances in Image-to-Video (I2V) generation allow a single image to be animated into a convincing video under text guidance, raising serious copyright and privacy risks.
We propose \method, an image protection approach that injects imperceptible perturbations into an image, inducing visible inconsistencies and structural failures in text-guided I2V generation.
Our method is motivated by a simple empirical observation: when text guidance is removed from modern I2V models, generation quality degrades markedly, not only in motion realism but also in subject preservation, structural coherence, and temporal consistency.
Building on this insight, \method exploits the model’s reliance on textual guidance by attenuating text-conditioned interactions during denoising while strengthening visual-only pathways.
To further systematically evaluate protection effectiveness, we introduce a Video-LLM–assisted evaluation protocol that provides interpretable, frame-grounded analyses of generation artifacts and inconsistencies.
Experiments on two representative I2V architectures demonstrate that our method achieves strong protection performance while improving efficiency and cross-model transferability.
\keywords{Safety and Privacy \and Image Protection \and I2V Generation}
\end{abstract}
\enlargethispage{\baselineskip}

\section{Introduction}
\label{sec:intro}

Diffusion models have rapidly advanced image synthesis~\cite{ddpm, ddim, dit, dalle2, imagen, sdxl, pixartalpha, flux, ldm, sd3} and video generation~\cite{vdm, imagenvideo, makeavideo, phenaki, lvdm, videoldm, cog, magi, ltx, wan, svd}, making it increasingly easy to animate a single photograph into a realistic short video.
Modern Image-to-Video (I2V) models~\cite{i2vgenxl, dynamicrafter, videocrafter1, videocomposer, svd} can generate temporally coherent motion and high visual fidelity from one image, guided by a text prompt.
These capabilities enable compelling content creation, but they also introduce a practical risk for publicly shared images~\cite{deepfakes_survey, mirsky2020deepfakes, bommasani2021foundation, weidinger2023sociotechnical}.
Once an image is posted online, it becomes a reusable source that can be repeatedly animated under a wide range of prompts, without the owner’s intent.
Compared to static misuse, video generation can be more persuasive because it adds actions, expressions, and temporal narratives, thereby amplifying privacy and copyright risks~\cite{deepfakes_survey}.
This motivates proactive image protection~\cite{photoguard, glaze, diffusionguard, prime, i2vguard} tailored to diffusion-based I2V dynamics.

In this work, we study I2V image protection.
Given an image intended for sharing, the owner adds an imperceptible perturbation so that attempts to animate the image are less likely to produce a coherent and image-faithful video.
Here, image-faithful refers to preserving the input image’s subject identity and overall structure throughout the generated video.
A successful protection, therefore, causes generated videos to exhibit noticeable inconsistencies or artifacts, making them unsuitable for convincing misuse.
Figure~\ref{fig:teaser} illustrates representative examples of such protected generations.

\begin{figure}
\centering
    \includegraphics[width=\linewidth]{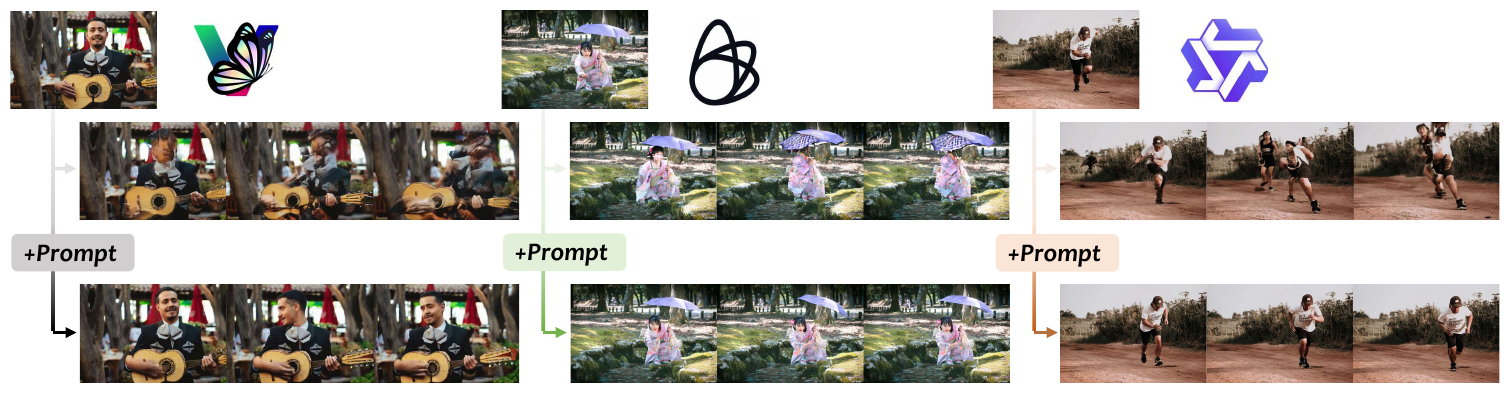}
    \caption{
\textbf{Empirical observation: prompt removal degrades I2V generation broadly.}
Top: prompt omitted (empty text). Bottom: standard text-guided generation with all other settings fixed.
Removing text guidance causes failures that extend beyond motion quality to subject preservation, structural coherence, and temporal consistency.
This pattern is observed across CogVideoX~\cite{cog}, LTX-Video~\cite{ltx}, and Wan~\cite{wan}.
}
\label{fig:cfg_motivation}
\end{figure}

Our approach is motivated by a simple empirical observation.
In several modern I2V models~\cite{cog,ltx,wan}, removing the text prompt during generation causes severe degradation that extends beyond prompt-following.
As illustrated in Figure~\ref{fig:cfg_motivation}, prompt-free generation often exhibits subject deformation, structural collapse, and temporal incoherence.
These failures are broad in scope: they affect not only motion quality but also subject preservation and structural consistency.
This behavior suggests that text guidance plays a broader role in stabilizing generation than typically assumed.

Based on this insight, we propose \method, an image protection method that weakens the influence of text during I2V denoising.
Rather than targeting specific prompts, our method targets the architectural pathways through which text affects generation.
Concretely, our objective suppresses text-conditioned interactions while reinforcing visual-only pathways, shifting the model’s updates away from text guidance.
We instantiate this idea for two common I2V architectures: full-attention models, where video tokens jointly attend to video, image, and text tokens, and cross-attention models, where text influences the latent through a dedicated residual branch.
Our optimization operates on intermediate attention statistics rather than generated videos, eliminating the need for expensive reference video generation and significantly improving efficiency compared with prior approaches~\cite{i2vguard}.

Finally, we introduce an evaluation protocol tailored to I2V protection.
Standard video benchmarks (\eg, VBench~\cite{vbench,vbench++,vbench2}) measure overall video quality but provide limited insight into how generation degrades on protected images.
In protection settings, success is inherently asymmetric: even a single salient error can render a generated video unsuitable for convincing misuse.
To better capture such behaviors, we complement standard benchmarks with a Video-LLM–assisted evaluation protocol~\cite{videochatgpt, videollama, videollava, qwen3vl, videollama3}.
The protocol evaluates four interpretable dimensions, \emph{Subject Preservation, Structural Consistency, Dynamic Consistency, and Artifact Suppression}, using structured outputs that cite frame-grounded observations before assigning discrete scores.

Our contributions are summarized as follows:
\begin{itemize}
\item We propose \method, an I2V image protection method that attenuates text-conditioned interactions during denoising while amplifying visual-only pathways, instantiated for both full-attention and cross-attention I2V architectures.
\item We introduce a Video-LLM-assisted evaluation protocol that provides inspectable, frame-grounded diagnoses of protection-induced failures across four visual dimensions.
\item Our method achieves effective disruption of plausible animation with improved efficiency and high imperceptibility across two representative I2V architectures under our evaluation suite.
\end{itemize}
\section{Related Work}
\label{sec:related}

\paragraph{\textbf{\textup{Image-to-Video Generative Models.}}}
Modern image-to-video (I2V) synthesis is largely built by extending text-to-image diffusion backbones with temporal modules to synthesize temporally coherent frames from a single image.
Early works such as Tune-A-Video~\cite{tav} and AnimateDiff~\cite{anidiff} demonstrated that injecting lightweight temporal components enables compelling image animation.
Stable Video Diffusion~\cite{svd} further scaled latent video diffusion and has served as a widely used open-source baseline.
Recent models improve fidelity and controllability through stronger spatio-temporal architectures, including Lumiere~\cite{lumiere} and large-scale diffusion transformers such as CogVideoX~\cite{cog}.
In parallel, LTX-Video~\cite{ltx} proposes efficiency-oriented designs and Wan~\cite{wan} continues to reduce the barrier to high-quality I2V generation as a large-scale open source model.
Community-driven efforts such as VideoCrafter2~\cite{videocrafter2} and Open-Sora~\cite{opensora} further broaden access to I2V capabilities.
The increasing realism and accessibility of I2V generation have spurred interest in protecting images against video synthesis.

\paragraph{\textbf{\textup{Image Protection against Generative Models.}}}
Protective perturbations aim to reduce the utility of images for downstream generation or editing by disrupting key components such as latent encoders, denoising dynamics, or attention.
PhotoGuard~\cite{photoguard} introduced diffusion-targeted image immunization to prevent unauthorized edits, and Distraction~\cite{distraction} improved practicality by focusing on attention-level objectives with reduced memory cost.
DiffusionGuard~\cite{diffusionguard} strengthens robustness under challenging masked-editing settings, while AdvPaint~\cite{advpaint} explicitly targets attention interactions to defend against inpainting manipulation.
Beyond editing, Glaze~\cite{glaze} and AdvDM~\cite{advdm} protect against artistic style imitation, and PRIME~\cite{prime} extends perturbation-based protection to video content.
Most closely related to our setting, I2VGuard~\cite{i2vguard} studies safeguards against diffusion-based I2V generation by degrading spatio-temporal consistency, but its optimization can require generating auxiliary videos and additional model passes.
These observations motivate more efficient I2V protection mechanisms that directly target text-dependent interactions to induce generation failures.
\section{Background}
\label{sec:pre}
\paragraph{\textbf{\textup{Diffusion Models.}}}
Diffusion models are trained to reverse a noise process that gradually corrupts data.
A noisy latent $z_t$ at the timestep $t$ is expressed as:
\begin{equation}
z_t = \sqrt{\bar{\alpha}_t}\, z_0 + \sqrt{1 - \bar{\alpha}_t}\, \epsilon,\quad \epsilon \sim \mathcal{N}(0, I),
\end{equation}
where $z_0$ is the clean latent and $\bar{\alpha}_t$ is the noise schedule.
A denoising network $\epsilon_\theta$ (\eg, U-Net~\cite{unet}, DiT~\cite{dit}) is trained to predict $\epsilon$ from $z_t$ at timestep $t$, under conditioning input $c$ (\eg, text embeddings).
In diffusion-based I2V systems, such conditioning is typically injected through attention mechanisms that control how signals from different modalities are combined~\cite{vaswani2017attention, ldm}.

\paragraph{\textbf{\textup{Conditioning Mechanisms in Diffusion Models.}}}
We focus on two representative conditioning designs commonly used in diffusion-based I2V models: \textit{Full-Attention} and \textit{Cross-Attention}.
To describe both architectures with a unified perspective, we distinguish \emph{text-dependent} interactions that explicitly access text tokens, from \emph{visual-only} interactions that operate on image and video tokens only.

\paragraph{\textbf{\textup{Full-Attention.}}}
In \textit{Full-Attention} architectures (\eg, CogVideoX~\cite{cog}), video tokens jointly attend to video, image, and text tokens under shared softmax normalization.
Under a shared softmax, modalities compete for attention mass~\cite{vaswani2017attention}, which directly determines their relative contribution to the latent update.
\begin{gather}
[\mathcal{A}_{z,z}, \mathcal{A}_{z,i}, \mathcal{A}_{z,\text{txt}}]
= \text{softmax}([A_{z,z}, A_{z,i}, A_{z,\text{txt}}]) \\
z^{\ell}_{FA}
= \mathcal{A}_{z,z}V_z^\ell + \mathcal{A}_{z,i}V_i^\ell + \mathcal{A}_{z,\text{txt}}V_\text{txt}^\ell,
\end{gather}
where $\ell$ is the layer index.
Here, $A_{z,\cdot}$ denotes the pre-softmax attention logits grouped by modality, $\mathcal{A}_{z,\cdot}$ the corresponding post-softmax weights, and $V_\cdot^\ell$ the value projections.
$z^{\ell}_{FA}$ denotes the output of the full-attention block.
Under our criterion, the term $\mathcal{A}_{z,\text{txt}}V_{\text{txt}}^\ell$ is \emph{text-dependent}, while $\mathcal{A}_{z,z}V_z^\ell + \mathcal{A}_{z,i}V_i^\ell$ constitutes \emph{visual-only} contributions.

\paragraph{\textbf{\textup{Cross-Attention.}}}
In \textit{Cross-Attention} architectures (\eg, LTX-Video~\cite{ltx}), text enters through a dedicated cross-attention residual, while self-attention operates on image and video tokens:
\begin{align}
    z^{\ell}_{SA} &= SA^{\ell}([z^\ell_i, z^{\ell}]) \label{eq:zsa}\\
    z^{\ell}_{CA} &= CA^{\ell}([z^\ell_i, z^{\ell}] + z^{\ell}_{SA}, c_\text{txt}) \label{eq:zca}\\
    [z_i^{\ell+1},z^{\ell+1}] &= [z^\ell_i, z^{\ell}] + z^{\ell}_{SA} + z^{\ell}_{CA},
\end{align}
where $z^{\ell}$ and $z^{\ell}_i$ denote the video latent and the image-condition latent in the $\ell$-th layer, $c_\text{txt}$ is the text condition, and $SA$ and $CA$ denote the self-attention and cross-attention modules, respectively.
In this design, text affects the video latent only through the cross-attention residual $z^{\ell}_{CA}$.
Thus, $z^{\ell}_{CA}$ is \emph{text-dependent}, whereas $z^{\ell}_{SA}$ and the image-conditioned pathway are \emph{visual-only}.
\section{Method}
\label{sec:method}
In this section, we describe our I2V image protection approach.
Our method consists of (1) the proposed \method objective that attenuates text influence during denoising, (2) an encoder attack that further distorts image conditioning, and (3) the overall objective to optimize the perturbation.

\begin{figure}
\centering
    \includegraphics[width=\linewidth]{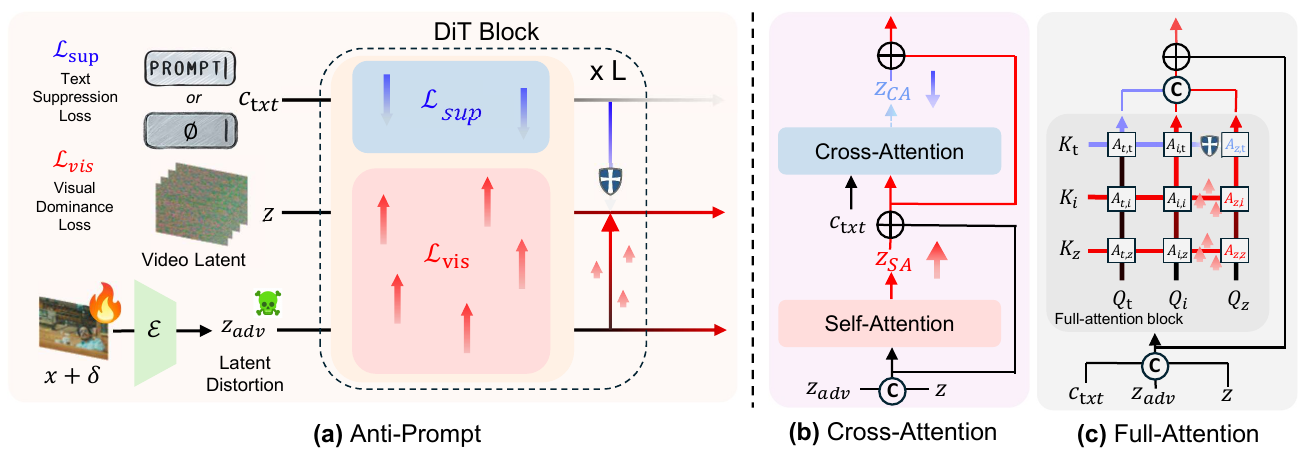}
    \caption{
\textbf{Overview of \method.}
(a) We optimize an imperceptible perturbation to suppress text-dependent pathways and promote visual-only dominance. Latent distortion corresponds to the encoder attack.
(b) \textbf{Cross-Attention:} text enters via a cross-attention residual. We suppress this text-dependent residual and strengthen the self-attention (visual-only) residual.
(c) \textbf{Full-Attention:} video tokens attend jointly to video, image, and text tokens under a shared softmax. We reduce video-to-text interactions and reinforce visual interactions to promote visual dominance.
}   
    \label{fig/framework}
\end{figure}

\subsection{\method Objective}
\label{sec:main_method}
The goal of I2V image protection is to induce salient failures in text-guided I2V generation from a protected image.
As shown in Section~\ref{sec:intro}, we observe that removing text prompts from I2V models produces broad generation failures not only in motion quality but also in subject preservation and structural coherence (Figure~\ref{fig:cfg_motivation}).
\method exploits this sensitivity: it optimizes an adversarial perturbation that attenuates text-conditioned interactions during denoising, reproducing, and intensifying the failure patterns seen under prompt removal.

\paragraph{\textbf{\textup{Text Suppression.}}}
Text suppression denotes attenuating text-dependent interactions during denoising.
Since text influence in diffusion models is mediated through attention interactions, these interactions provide a natural target for disruption.
We suppress text influence by penalizing direct text-dependent interactions at each layer:
\begin{equation}
\mathcal{L}_{\text{sup}} = \sum_{\ell=1}^{L} \tau^{\ell}.
\label{eq:suppression}
\end{equation}
Here, $\tau^\ell$ is a scalar that measures the strength of text-dependent pathways at layer $\ell$.
For \textit{Full-Attention}, we set $\tau^\ell=\mathbb{E}[A^{\ell}_{z,\text{txt}}]$, \ie, the average pre-softmax attention logits from video to text tokens, where $\mathbb{E}[\cdot]$ denotes averaging over all attention heads and the corresponding token pairs.
We suppress the text-related logits $A^{\ell}_{z,\text{txt}}$ to attenuate text-conditioned pathways.
For \textit{Cross-Attention}, we set $\tau^\ell=\mathbb{E}[\|z^{\ell}_{CA}\|_2^2]$, \ie, the energy of the cross-attention residual that injects text features into the video latent, thereby shrinking text-dependent residual updates.

\paragraph{\textbf{\textup{Visual Dominance.}}}
Text suppression directly weakens text-conditioned pathways, but its effect depends on the specific text tokens encountered during optimization and may transfer less reliably to unseen prompts.
Visual dominance complements this by amplifying visual-only interactions regardless of prompt content, reducing the relative contribution of text guidance.
By increasing the magnitude of activations that exclude text, we reduce textual influence through softmax competition in \textit{Full-Attention} and residual dominance in \textit{Cross-Attention}:
\begin{equation}
\mathcal{L}_{\text{vis}} = -\sum_{\ell=1}^{L} \nu^{\ell}.
\label{eq:dominance}
\end{equation}
Here, $\nu^\ell$ is a scalar that measures the strength of visual-only pathways at layer $\ell$.
For \textit{Full-Attention}, we set $\nu^\ell=\mathbb{E}[A^{\ell}_{z,z}]+\mathbb{E}[A^{\ell}_{z,i}]$, \ie, the average pre-softmax attention logits from video to video and image tokens.
Since these modalities compete under the shared softmax, increasing $\nu^\ell$ makes the visual terms more competitive and consequently decreases the relative contribution of text guidance.
For \textit{Cross-Attention}, we set $\nu^\ell=\mathbb{E}[\|z^{\ell}_{SA}\|_2^2]$, \ie, the energy of the self-attention residual, encouraging it to dominate residual updates and reducing the relative contribution of cross-attention text features.

These complementary strategies provide an intuitive realization of our protection objective.
Although the formulation differs for \textit{Full-Attention} and \textit{Cross-Attention}, both are designed to attenuate text-guided control and drive denoising toward failures that persist across unseen prompts.

\subsection{Encoder Attack}
Following~\cite{i2vguard}, we adopt an encoder attack~\cite{photoguard} to further weaken image conditioning.
Empirically, while prompt omission often yields unstable outputs, some generations can still retain partial coherence depending on the pipeline and scene.
To further reduce such residual image-conditioned coherence, we drive the encoder representation toward an uninformative target.
Specifically, we perturb the input so that $\mathcal{E}(x+\delta)$ approaches $\mathcal{E}(x_{\mathrm{tar}})$ (\eg, a black image), suppressing usable visual cues for generation:
\begin{equation}
\mathcal{L}_{\text{enc}} = \| \mathcal{E}(x+\delta) - \mathcal{E}(x_{\mathrm{tar}}) \|_2^2,
\label{eq:objective_enc}
\end{equation}
where $x$ is the input image, $\delta$ the perturbation, $\mathcal{E}$ the VAE encoder, and $x_{\mathrm{tar}}$ the uninformative target.

\subsection{Overall Loss}
To create the final protected image, we obtain the optimal imperceptible perturbation $\delta$ by jointly optimizing all three loss components.
The final objective is formulated as:
\begin{align}
\min_{\delta} \mathcal{L} &= \mathcal{L}_{\text{sup}} + \lambda_1 \mathcal{L}_{\text{vis}} + \lambda_2 \mathcal{L}_{\text{enc}}, \label{eq:final} \\ 
&\text{s.t.} \quad \| \delta \|_{\infty} \leq \epsilon, \nonumber
\end{align}
where $\lambda_1$ and $\lambda_2$ are balancing hyperparameters, and the constraint ensures imperceptibility.
By jointly optimizing these objectives, our method learns a perturbation that reduces text-conditioned influence and makes text-guided generation more likely to exhibit visible failures.
An overview of the proposed \method framework and its detailed optimization procedure is illustrated in Figure~\ref{fig/framework}.

\section{Evaluation Protocol for Text-Guided I2V Protection}
\label{sec:eval_proto}
Evaluating I2V image protection is fundamentally different from evaluating video generation quality.
Prior I2V protection work~\cite{i2vguard} evaluates performance using video generation benchmarks such as VBench I2V~\cite{vbench, vbench++} designed to assess how well a generated video satisfies multiple perceptual criteria.
In contrast, the objective of protection is to break image-faithful animation of the protected input.
Accordingly, protection effectiveness can often be established by identifying a single clear failure (\eg, subject drift, structural breakdown, temporal instability, and prominent artifacts) rather than by exhaustively scoring fine-grained quality.

Motivated by this distinction, we introduce a failure-finding evaluation protocol that characterizes how generated videos depart from an image-faithful animation.
Our protocol scores videos along four interpretable visual dimensions that capture common I2V failure patterns.
Each dimension is rated on a discrete 1--5 scale, where higher scores indicate more stable, coherent generation and lower scores indicate more severe failures.
Thus, lower scores correspond to stronger protection.

\subsection{Evaluation Dimensions}
We score the following dimensions because protection can disrupt I2V generation through distinct mechanisms, and mixing them obscures diagnosis.
Each dimension is scored on a discrete 1--5 scale, where \emph{5} indicates that the dimension is largely preserved and \emph{1} indicates a severe failure.
Accordingly, \emph{lower scores indicate stronger protection}.

\paragraph{Subject Preservation.}
We assess whether the main subject from the first frame remains present and recognizable across the video.
This captures identity drift, disappearance, and replacement, which are often the most salient failures even when the overall rendering still looks plausible.

\paragraph{Structural Consistency.}
We examine whether the subject's geometry and boundaries remain stable across frames.
This highlights warping, tearing, and part-level inconsistencies that can appear even when the subject is still identifiable.

\paragraph{Dynamic Consistency.}
We check whether motion and state transitions remain temporally coherent and physically plausible.
This targets flicker, abrupt jumps, and implausible dynamics that may occur without large appearance changes.

\paragraph{Artifact Suppression.}
We assess how well visible artifacts are suppressed (\eg, checkerboard patterns, banding, flickering textures, unnatural noise, or rendering glitches).
Higher scores indicate minimal visible artifacts, whereas lower scores indicate severe and dominant artifacts.

\begin{table}[t]
\scriptsize
\centering
\caption{White-box quantitative results on VBench metrics~\cite{vbench,vbench++}. Lower indicates stronger protection efficacy, since our metrics are formulated to quantify deviations from an image-faithful animation.}
\resizebox{\linewidth}{!}{
\begin{tabular}{l cccccccc}
\toprule
{Method} &
\makecell{I2V\\Subject} &
\makecell{I2V\\Background} &
\makecell{Subject\\Consistency} &
\makecell{Background\\Consistency} &
\makecell{Aesthetic\\Quality} &
\makecell{Imaging\\Quality} &
\makecell{Temporal\\Flickering} &
\makecell{Motion\\Smoothness} \\
\midrule
\multicolumn{9}{l}{\textcolor{gray}{\textit{CogVideoX~\cite{cog}}}} \\
Clean & 97.15 & 98.87 & 95.06 & 97.38 & 61.48 & 70.23 & 96.99 & 98.15 \\
I2VGuard &  94.58 & 97.14 & 93.32 & 96.17 & 58.61 & 65.30 & 96.47 & 97.61 \\
Ours & \textbf{91.20} & \textbf{94.28} & \textbf{87.54} & \textbf{93.16} & \textbf{54.72} & \textbf{63.71} & \textbf{94.47} & \textbf{95.66} \\
\hline
\multicolumn{9}{l}{\textcolor{gray}{\textit{LTX-Video~\cite{ltx}}}} \\
Clean & 97.70 & 98.15 & 95.09 & 97.87 & 60.87 & 68.22 & 99.04 & 99.36 \\
I2VGuard  & 94.36 & 95.56 & 90.25 & 95.00 & 56.28 & 62.63 & 98.23 & 98.90 \\
Ours  & \textbf{93.64} & \textbf{95.07} & \textbf{89.36} & \textbf{94.99} & \textbf{54.99} & \textbf{61.44} & \textbf{97.95} & \textbf{98.67} \\
\bottomrule
\end{tabular}
}
\label{tab:vbench}
\end{table}

\begin{table}[t]
\scriptsize
\centering
\caption{VBench~\cite{vbench,vbench++} results of generated videos with unseen prompt.}
\resizebox{\linewidth}{!}{
\begin{tabular}{l cccccccc}
\toprule
{Method} &
\makecell{I2V\\Subject} &
\makecell{I2V\\Background} &
\makecell{Subject\\Consistency} &
\makecell{Background\\Consistency} &
\makecell{Aesthetic\\Quality} &
\makecell{Imaging\\Quality} &
\makecell{Temporal\\Flickering} &
\makecell{Motion\\Smoothness} \\
\midrule
\multicolumn{9}{l}{\textcolor{gray}{\textit{CogVideoX~\cite{cog}}}} \\
Clean & 96.87 & 97.65 & 94.88 & 97.40 & 60.84 & 70.21 & 97.41 & 98.31 \\
I2VGuard &  93.51 & 93.79 & 92.45 & 95.19 & 57.37 & 63.58 & 96.34 & 97.43 \\
Ours & \textbf{88.90} & \textbf{90.60} & \textbf{84.70} & \textbf{91.76} & \textbf{53.14} & \textbf{61.80} & \textbf{94.10} & \textbf{95.36} \\
\hline
\multicolumn{9}{l}{\textcolor{gray}{\textit{LTX-Video~\cite{ltx}}}} \\
Clean & 95.85 & 97.30 & 90.12 & 95.50 & 56.96 & 64.37 & 98.53 & 99.11 \\
I2VGuard  & 91.83 & 92.09 & 83.46 & 90.28 & 50.69 & 56.36 & 97.35 & 98.40 \\
Ours  & \textbf{91.14} & \textbf{91.38} & \textbf{82.47} & \textbf{90.14} & \textbf{49.92} & \textbf{55.85} & \textbf{96.93} & \textbf{98.06} \\
\bottomrule
\end{tabular}
}
\label{tab:unseen_prompt}
\end{table}

\subsection{Video-LLM scoring with Inspectable Evidence}
For each generated video, we uniformly sample 4 frames per second, including the first and last.
Given the sampled frames, a Video-LLM produces (1) a small set of concrete observations grounded to frame indices, and then (2) a 1--5 score for each dimension.
Requiring observations before scoring makes the output more inspectable and reduces arbitrary ratings~\cite{zheng2023judging, liu2023geval}.

To evaluate videos at scale under this protocol, we employ Video-LLMs~\cite{qwen3vl, videollama3, videochatgpt, videollama, videollava} as evaluators that inspect sampled frames from each generated video and assign dimension-wise scores based on observable visual evidence.
The prompt requests a structured output containing frame-grounded observations and per-dimension scores.
We release the prompt template and parsing scripts to facilitate reproducibility.
We report both the per-dimension scores and their average as a compact summary statistic.

Video-LLM--based evaluation is inherently imperfect and may exhibit model-specific biases or stochastic variability.
We therefore complement it with standard VBench metrics and validate the reliability through a human study in Section~\ref{sec:exp}.
As shown in Table~\ref{tab:human_study}, the human judgments align well with our protocol-induced rankings, supporting its use as a scalable failure-finding evaluation.
Additional evaluator repeatability and cross-evaluator consistency analyses are provided in the supplementary material.
\section{Experiments}
\label{sec:exp}
\begin{figure*}[t]
\centering
    \includegraphics[width=\linewidth]{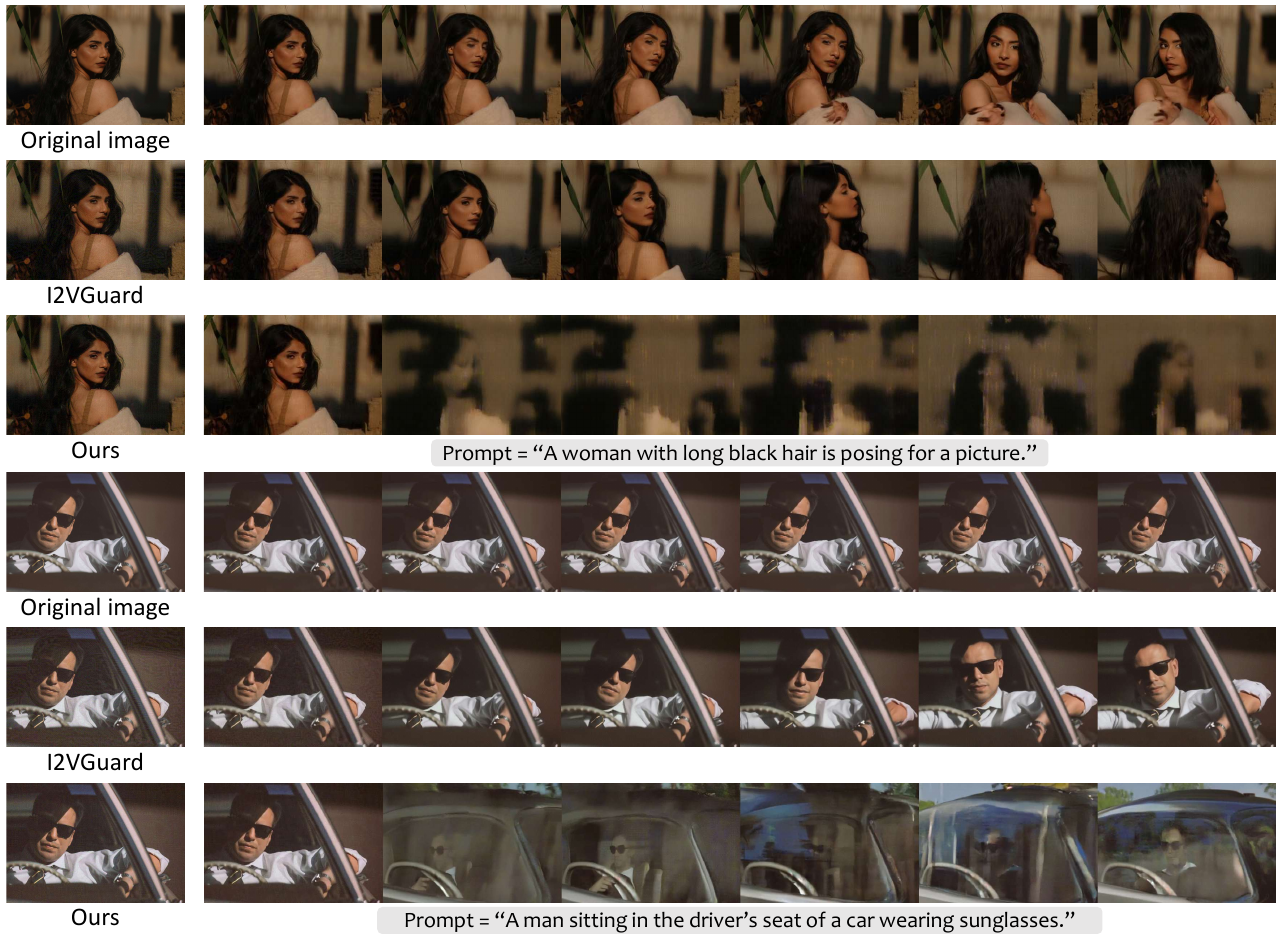}
\caption{\textbf{Qualitative results on CogVideoX~\cite{cog}.}
We compare generations from the original image, I2VGuard~\cite{i2vguard}, and Ours.}
\label{fig:cog_qual}
\end{figure*}

\begin{figure*}[t]
\centering
\includegraphics[width=\linewidth]{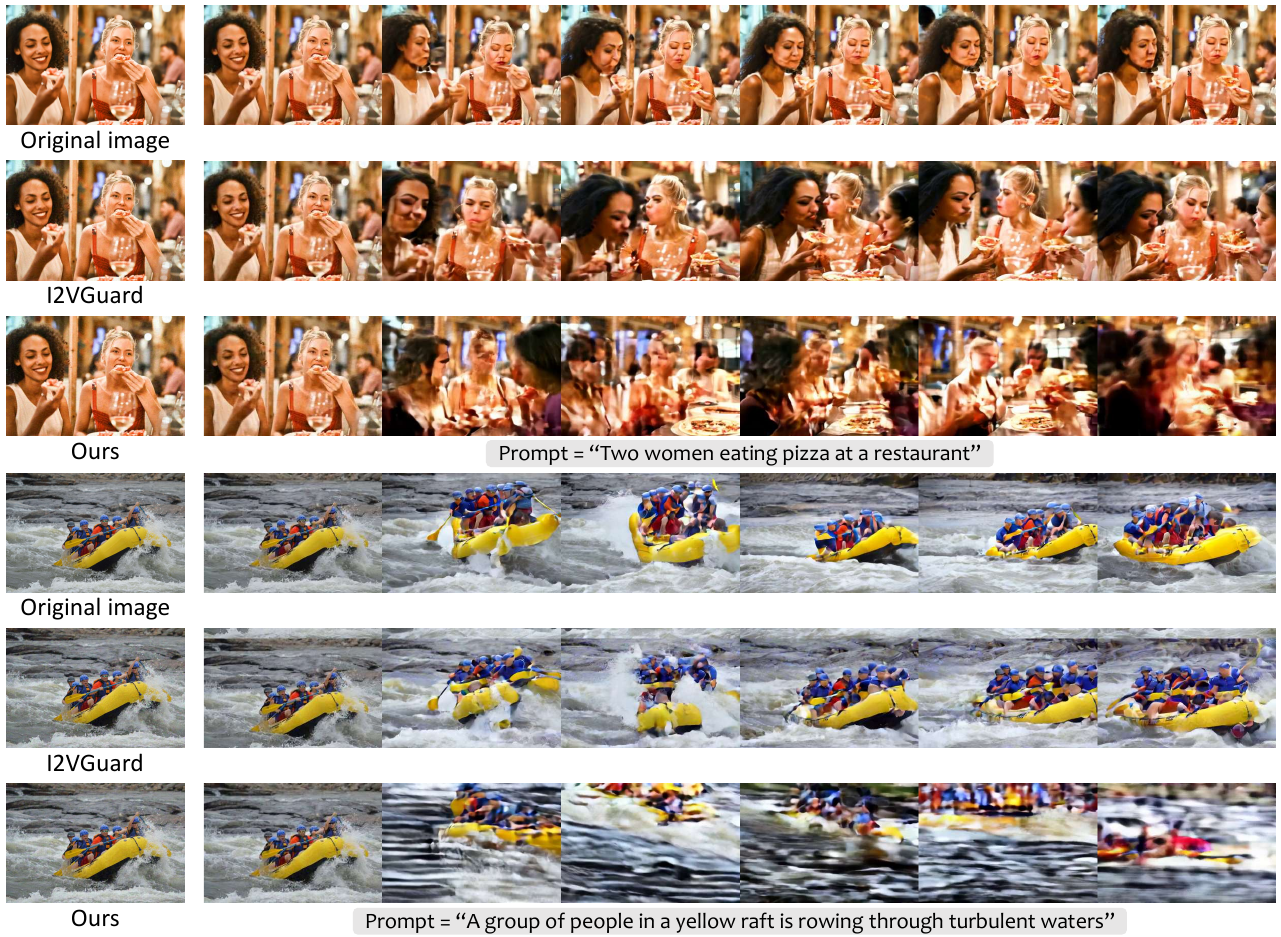}
\caption{\textbf{Qualitative results on LTX-Video~\cite{ltx}.}
We compare generations from the original image, I2VGuard~\cite{i2vguard}, and Ours.}
\label{fig:ltx_qual}
\end{figure*}

\paragraph{\textbf{\textup{Implementation Details.}}}
Following I2VGuard~\cite{i2vguard}, we set the PGD hyperparameters to $K=50$, $\epsilon=8/255$, and $\alpha=1/255$ for all experiments.
We unroll $T=10$ denoising steps during optimization.
Unless specified otherwise, we set $\lambda_1=1$ and $\lambda_2=1$ for CogVideoX~\cite{cog}, and use $\lambda_1=0.1$ and $\lambda_2=0.01$ for LTX-Video~\cite{ltx} to account for objective scale differences across architectures.
For video generation, we use $720\times480$ resolution with 49 frames at 8 FPS for CogVideoX, and $704\times480$ with 121 frames at 25 FPS for LTX-Video.

\paragraph{\textbf{\textup{Experimental Setup.}}}
We employ representative open-source image-to-video generators in our text-guided I2V setting.
Specifically, we use CogVideoX-5B~\cite{cog} (full-attention) and LTX-Video-2B~\cite{ltx} (cross-attention) as target models, as they cover the two common conditioning designs and remain feasible to run at scale under practical memory budgets.

\begin{table}[t]
\scriptsize
\centering
\caption{Black-box transfer evaluation on VBench metrics~\cite{vbench,vbench++}.}
\resizebox{\linewidth}{!}{
\begin{tabular}{l cccccccc}
\toprule
{Method} &
\makecell{I2V\\Subject} &
\makecell{I2V\\Background} &
\makecell{Subject\\Consistency} &
\makecell{Background\\Consistency} &
\makecell{Aesthetic\\Quality} &
\makecell{Imaging\\Quality} &
\makecell{Temporal\\Flickering} &
\makecell{Motion\\Smoothness} \\
\midrule
\multicolumn{9}{l}{\textcolor{gray}{\textit{Optimization: LTX-Video~\cite{ltx}, Generation: CogVideoX~\cite{cog}}}} \\
Clean & 97.15 & 98.87 & 95.06 & 97.38 & 61.48 & 70.23 & 96.99 & 98.15 \\
I2VGuard & 97.49 & 98.62 & 95.41 & 97.31 & 60.77 & 70.43 & 96.98 & 98.24 \\
Ours & \textbf{95.28} & \textbf{96.72} & \textbf{94.34} & \textbf{96.57} & \textbf{59.56} & \textbf{68.42} & \textbf{96.56} & \textbf{97.83} \\
\hline
\multicolumn{9}{l}{\textcolor{gray}{\textit{Optimization: CogVideoX~\cite{cog}, Generation: LTX-Video~\cite{ltx}}}} \\
Clean & 97.70 & 98.15 & 95.09 & 97.87 & 60.87 & 68.22 & 99.04 & 99.36 \\
I2VGuard  & \textbf{94.52} & \textbf{95.33} & 90.98 & 95.33 & 55.78 & \textbf{61.42} & 98.29 & 98.90 \\
Ours  & 94.61 & 95.60 & \textbf{89.87} & \textbf{94.98} & \textbf{55.40} & 62.63 & \textbf{98.14} & \textbf{98.83} \\
\bottomrule
\end{tabular}
}
\label{tab:blackbox_vbench}
\end{table}

\paragraph{\textbf{\textup{Datasets and Metrics.}}}
We evaluate on 355 images from the VBench Image-to-Video dataset~\cite{vbench++}, using the ground-truth text prompts paired with each image for consistent evaluation.
We report VBench~\cite{vbench,vbench++} metrics and our proposed failure-finding scores computed with an open-source Video-LLM evaluator (Qwen3-VL-8B~\cite{qwen3vl}).
We follow the official VBench implementation for metric computation; our failure-oriented evaluation protocol is described in Section~\ref{sec:eval_proto}.

\begin{table}[t]
\scriptsize
\setlength{\tabcolsep}{4pt}
\centering
\caption{White-box quantitative results under our I2V protection evaluation protocol. Lower scores indicate more severe failures. Higher scores indicate more faithful animation of the input image. Scores are reported as \textit{a/b} for \textit{seen/unseen} prompts.}
\resizebox{0.85\linewidth}{!}{
\begin{tabular}{@{}lccccc@{}}
\toprule
Method
& \makecell{Subject\\Preservation}
& \makecell{Structural\\Consistency}
& \makecell{Dynamic\\Consistency}
& \makecell{Artifact\\Suppression}
& \makecell{Average} \\
\midrule
\multicolumn{6}{@{}l}{\textcolor{gray}{\textit{CogVideoX~\cite{cog}}}}\\
% \multicolumn{6}{@{}l}{\textcolor{gray}{\textit{CogVideoX}}}\\
Clean    & 4.86/4.72 & 4.55/4.22 & 4.19/3.91 & 4.79/4.66 & 4.58/4.38 \\
I2VGuard & 4.72/4.71 & 4.18/3.95 & 3.81/3.66 & 2.97/3.13 & 3.92/3.86 \\
Ours     & \textbf{4.37}/\textbf{4.26} & \textbf{3.24}/\textbf{2.82} & \textbf{2.91}/\textbf{2.61} & \textbf{2.49}/\textbf{2.28} & \textbf{3.25}/\textbf{2.99} \\
\midrule
\multicolumn{6}{@{}l}{\textcolor{gray}{\textit{LTX-Video~\cite{ltx}}}}\\
% \multicolumn{6}{@{}l}{\textcolor{gray}{\textit{LTX-Video}}}\\
Clean    & 4.64/3.78 & 4.01/2.83 & 3.70/2.85 & 4.17/3.08 & 4.13/3.13 \\
I2VGuard & 4.24/3.37 & 3.18/2.21 & 2.99/\textbf{2.17} & 2.70/1.95 & 3.28/2.46 \\
Ours     & \textbf{4.06}/\textbf{3.21} & \textbf{3.02}/\textbf{2.20} & \textbf{2.85}/\textbf{2.17} & \textbf{2.58}/\textbf{1.92} & \textbf{3.12}/\textbf{2.38} \\
\bottomrule
\end{tabular}
}
\label{tab:wb_cog}
\end{table}

\begin{table}[t]
\scriptsize
\setlength{\tabcolsep}{4pt}
\centering
\caption{Black-box transfer evaluation on our evaluation protocol.}
\resizebox{0.82\linewidth}{!}{
\begin{tabular}{@{}lccccc@{}}
\toprule
Method
& \makecell{Subject\\Preservation}
& \makecell{Structural\\Consistency}
& \makecell{Dynamic\\Consistency}
& \makecell{Artifact\\Suppression}
& \makecell{Average} \\
\midrule
\multicolumn{6}{@{}l}{\textcolor{gray}{\textit{Optimization: LTX-Video~\cite{ltx}, Generation: CogVideoX~\cite{cog}}}} \\
Clean    & 4.86 & 4.55 & 4.19 & 4.79 & 4.58 \\
I2VGuard & \textbf{4.84} & 4.44 & 4.09 & 4.50 & 4.47 \\
Ours     & \textbf{4.84} & \textbf{4.31} & \textbf{3.92} & \textbf{3.51} & \textbf{4.15} \\
\midrule
\multicolumn{6}{@{}l}{\textcolor{gray}{\textit{Optimization: CogVideoX~\cite{cog}, Generation: LTX-Video~\cite{ltx}}}}\\
Clean    & 4.64 & 4.01 & 3.70 & 4.17 & 4.13 \\
I2VGuard & 4.24 & 3.23 & 3.14 & 2.97 & 3.40 \\
Ours     & \textbf{4.07} & \textbf{3.11} & \textbf{2.95} & \textbf{2.86} & \textbf{3.25} \\
\bottomrule
\end{tabular}
}
\label{tab:blackbox_transfer}
\end{table}
\paragraph{\textbf{\textup{Baselines.}}}
We compare our method with I2VGuard~\cite{i2vguard}, which is designed to prevent unauthorized video generation.
The original I2VGuard paper reports results on a non-public dataset, so exact numeric matching is not possible.
We reproduce I2VGuard using the hyperparameters and design choices described in the paper, and evaluate all methods on the public VBench I2V benchmark under identical generation settings and perturbation budgets.
In our setup, the reproduced I2VGuard achieves effectiveness broadly consistent with the reported results.
For fair comparison, both methods use the encoder attack by default and share the same uninformative target (a black image).

\begin{table}[t]
\centering
\caption{Human study and protocol alignment.
The study uses a blinded triplet comparison over Clean, I2VGuard, and Ours. A higher worst rate indicates stronger protection-induced degradation.}
\resizebox{0.72\linewidth}{!}{
\begin{tabular}{lcccc}
\toprule
Method & Cons. Rank $\downarrow$ & Plaus. Rank $\downarrow$
& Best (\%) $\uparrow$ & Worst (\%) $\uparrow$ \\
\midrule
Clean & 1.50 & 1.57 & 59.5 & 10.0 \\
I2VGuard & 1.77 & 1.74 & 33.5 & 10.9 \\
Ours & 2.72 & 2.69 & 7.0 & 79.1 \\
\midrule
\multicolumn{5}{c}{Protocol--human ranking agreement} \\
\midrule
Criterion & \multicolumn{2}{c}{Spearman $\rho$}
& \multicolumn{2}{c}{Kendall $\tau$} \\
\midrule
Overall ranking & \multicolumn{2}{c}{0.84}
& \multicolumn{2}{c}{0.71} \\
Consistency & \multicolumn{2}{c}{0.81}
& \multicolumn{2}{c}{0.68} \\
Plausibility & \multicolumn{2}{c}{0.79}
& \multicolumn{2}{c}{0.66} \\
\bottomrule
\end{tabular}}
\label{tab:human_study}
\end{table}

\begin{table}[t]
\centering
\scriptsize
\caption{Imperceptibility comparison.}
\resizebox{0.6\linewidth}{!}{
\begin{tabular}{lccccc}
\toprule
\textbf{Model} & \textbf{Method} & DISTS $\downarrow$ & LPIPS $\downarrow$ & PSNR $\uparrow$ & SSIM $\uparrow$ \\
\midrule
\multirow{2}{*}{CogVideoX} 
& I2VGuard & 0.138 & 0.246 & 32.86 & 0.838 \\
& Ours     & \textbf{0.111} & \textbf{0.196} & \textbf{35.10} & \textbf{0.906} \\
\midrule
\multirow{2}{*}{LTX-Video} 
& I2VGuard & \textbf{0.130} & 0.237 & 32.35 & 0.816 \\
& Ours     & {0.138} & \textbf{0.228} & \textbf{34.20} & \textbf{0.892} \\
\bottomrule
\end{tabular}
}

\label{tab:imper}
\end{table}

\paragraph{\textbf{\textup{Qualitative Results}}}
We compare generations from the original image and protected images optimized by I2VGuard~\cite{i2vguard} and Ours.
Figures~\ref{fig:cog_qual} and~\ref{fig:ltx_qual} show that our method more reliably induces salient failures and temporally unstable dynamics than the baseline across both CogVideoX and LTX-Video.
Overall, videos generated from our protected images become incoherent and inconsistent across frames, while the input perturbations remain imperceptible.
These failures are consistent with the degradation patterns observed under prompt removal (Figure~\ref{fig:cfg_motivation}), further supporting our motivation of disrupting text-guided stabilization.

\subsection{Quantitative Evaluation}
We first evaluate on the standard VBench benchmark~\cite{vbench,vbench++}.
Throughout this section, we use a unified convention: lower scores indicate more severe failures (stronger protection), since our metrics are formulated to quantify deviations from an image-faithful animation.
As shown in Table~\ref{tab:vbench}, our approach improves protection effectiveness over I2VGuard across various I2V metrics.
We further evaluate using our failure-finding protocol with an open-source Video-LLM evaluator (Qwen3-VL-8B~\cite{qwen3vl} in our implementation).
As shown in Table~\ref{tab:wb_cog}, our method achieves consistently lower scores across the four dimensions, indicating stronger disruption of image-faithful generation.

\paragraph{\textbf{\textup{Robustness against Unseen Prompts.}}}
In real-world scenarios, the defender has no access to the attacker’s prompt, making generalization to unseen prompts critical for dependable protection.
Accordingly, we evaluate robustness under unseen generation prompts by using GPT-4~\cite{gpt4} to synthesize diverse, plausible video scenarios conditioned on each input image.
We note that the generated unseen prompts are typically more informative and dynamic than the original VBench~\cite{vbench} prompts, making generation itself more challenging.
As a result, absolute scores tend to degrade for all methods, and this effect is particularly pronounced for LTX-Video~\cite{ltx}.
Nevertheless, our method maintains a clear advantage over I2VGuard~\cite{i2vguard} in this harder regime, indicating that the learned perturbations transfer robustly even when the prompt distribution shifts (Table~\ref{tab:unseen_prompt} and~\ref{tab:wb_cog}).
We use the same unseen prompts for all methods and clean generations.
The unseen prompts are provided in the supplementary material.

\paragraph{\textbf{\textup{Black-box Transferability.}}}
We further conduct cross-architecture black-box experiments to assess protection transferability.
As shown in Table~\ref{tab:blackbox_vbench}, \method is generally favorable to I2VGuard~\cite{i2vguard} in black-box transfer.
When optimized on LTX-Video~\cite{ltx}, it outperforms I2VGuard across all reported VBench metrics, while the CogVideoX~\cite{cog}-to-LTX setting shows a few exceptions on isolated dimensions but still favors \method overall.
In Table~\ref{tab:blackbox_transfer}, \method exhibits a clearer and more consistent advantage under our evaluation protocol in Sec.~\ref{sec:eval_proto}.
Taken together, these results indicate effective black-box transfer for \method, with its advantage appearing more consistently under the proposed failure-finding evaluation.

\begin{table}[t]
\scriptsize
\centering
\caption{Ablation of \method components.}
\resizebox{\linewidth}{!}{
\begin{tabular}{l cccccccc !{\vrule width 0.7pt} c}
\toprule
{Method} &
\makecell{I2V\\Subject} &
\makecell{I2V\\Background} &
\makecell{Subject\\Consistency} &
\makecell{Background\\Consistency} &
\makecell{Aesthetic\\Quality} &
\makecell{Imaging\\Quality} &
\makecell{Temporal\\Flickering} &
\makecell{Motion\\Smoothness} &
\makecell{Video-LLM~\cite{qwen3vl}\\Average$\downarrow$} \\
\midrule
$\mathcal{L}_{\text{enc}}$ & 94.47 & 97.71 & 91.96 & 96.76 & 57.83 & 64.22 & 96.23 & 97.12 & 4.11 \\
$\mathcal{L}_{\text{enc}}+\mathcal{L}_{\text{sup}}$ & 94.77 & 97.22 & 92.94 & 96.22 & 58.33 & 66.55 & 96.57 & 97.53 & 3.92 \\
$\mathcal{L}_{\text{enc}}+\mathcal{L}_{\text{vis}}$ & 92.92 & 95.91 & 90.53 & 94.90 & 56.38 & 66.95 & 95.32 & 96.66 & 3.59  \\
\cdashline{1-10}[1pt/1pt]
Ours  & 91.20 & 94.28 & 87.54 & 93.16 & 54.72 & 63.71 & 94.47 & 95.66 & 3.25 \\
\bottomrule
\end{tabular}}
\label{tab:abl}
\end{table}

\begin{table}[t]
\centering
\scriptsize
\caption{Efficiency comparison.}
\resizebox{0.8\linewidth}{!}{
\begin{tabular}{l l c c c c}
\toprule
\textbf{Model} & \textbf{Method} & \makecell{Video Gen. \\ (PFLOPs)} & \makecell{$\delta$ Optim. \\ (PFLOPs)} & \makecell{Total FLOPs \\ (PFLOPs)} & \makecell{Peak Memory \\ (GB)} \\
\midrule
\multirow{2}{*}{CogVideoX-5B} 
& I2VGuard & 16.96 & 37.29 & 54.25 & 51.56 \\
& Ours     & \textbf{0} & \textbf{26.82} & \textbf{26.82} & \textbf{48.98} \\
\hline
\multirow{2}{*}{LTX-Video-2B} 
& I2VGuard & 1.896 & 3.517 & 5.413 & 35.18 \\
& Ours     & \textbf{0} & \textbf{2.609} & \textbf{2.609} & \textbf{28.86} \\
\bottomrule
\end{tabular}
}
\label{tab:efficiency}
\end{table}
\paragraph{\textbf{\textup{Human Study and Protocol Alignment.}}}
To verify that the generated failures are perceptually meaningful, we conduct a blinded human study on generated video triplets. Participants compare the results from Clean, I2VGuard, and Ours for the same image and prompt, and select the best and worst videos based on consistency and plausibility.
To reduce the cognitive burden of comparing video triplets, we ask participants to judge two higher-level criteria rather than all four protocol dimensions: consistency, which mainly reflects subject preservation and structural consistency, and plausibility, which mainly reflects dynamic consistency and artifact suppression.
As shown in Table~\ref{tab:human_study}, generations from our protected images are most frequently selected as the worst, indicating stronger human-perceived degradation.
The protocol-induced rankings also agree well with aggregated human preferences across overall, consistency, and plausibility criteria.

\subsection{Analysis on Imperceptibility}
Beyond protection effectiveness, practical protective perturbation should be visually imperceptible.
We measure DISTS~\cite{dists}, LPIPS~\cite{lpips}, PSNR, and SSIM~\cite{ssim}, comparing our method with I2VGuard~\cite{i2vguard}.
As shown in Table~\ref{tab:imper}, our perturbations achieve consistently lower LPIPS and higher PSNR/SSIM across both models, indicating improved perceptual fidelity.
DISTS is improved on CogVideoX and is slightly higher on LTX-Video.

\subsection{Ablation Studies}
\paragraph{\textbf{\textup{Impact of Each Proposed Component.}}}
We conduct ablation experiments to evaluate the contribution of our two objectives,
$\mathcal{L}_{\text{sup}}$ and $\mathcal{L}_{\text{vis}}$, as shown in Table~\ref{tab:abl}.
In this analysis, the encoder attack ($\mathcal{L}_\text{enc}$) is used as a baseline component.
We evaluate the protection efficacy when adding $\mathcal{L}_{\text{sup}}$ only, $\mathcal{L}_{\text{vis}}$ only, and both combined.
All ablation experiments are conducted in the white-box setting on CogVideoX~\cite{cog}.

Adding $\mathcal{L}_{\text{sup}}$ alone shows mixed effects on VBench metrics, where some scores slightly increase, but consistently improve the Video-LLM failure score.
This suggests that text suppression induces failures that are visible to a Video-LLM evaluator inspecting individual frames but are not fully captured by the automated VBench pipeline.
Adding $\mathcal{L}_{\text{vis}}$ alone produces consistent improvements across both evaluation frameworks.
Their combination achieves the lowest scores on both VBench and the Video-LLM protocol, confirming that text suppression and visual dominance are complementary.

\paragraph{\textbf{\textup{Efficiency Analysis.}}}
As shown in Table~\ref{tab:efficiency}, our method significantly reduces runtime and memory consumption compared to I2VGuard~\cite{i2vguard}.
This is because our losses operate on intermediate attention statistics within a single denoising pass, whereas I2VGuard requires generating a reference video as an additional forward pass.
This reduced memory footprint makes the optimization feasible under practical GPU budgets and leaves headroom to scale to larger I2V models when more compute becomes available.

\section{Conclusion}
\label{sec:conclusion}
We introduced \method, an image protection strategy that aims to prevent unauthorized text-guided image-to-video generation from publicly shared images.
\method \ combines text suppression to attenuate text-dependent interactions with a visual dominance loss that increases robustness to unseen prompts.
We further propose a Video-LLM-based failure-finding evaluation protocol that exposes protection-induced breakdowns along four visual dimensions.
Extensive experiments on VBench and our protocol across cross-attention and full-attention I2V models demonstrate strong protection with high imperceptibility and improved computational efficiency.
Moreover, our method demonstrates generally stronger protection efficacy in unseen-prompt and model-shift evaluation settings.

\clearpage
% \subsection*{Acknowledgement}
% This work was supported by the AI Graduate School Program at POSTECH (RS-2019-II191906 (5\%)), the NRF grants (RS-2025-24535146 (10\%), RS-2026-25491789 (40\%)), the IITP grants (RS-2022-II220926 (25\%)) funded by MSIT, and the KRIT grant (No. 21-107-E00-009-02 (20\%)) funded by DAPA, Korea.
% This work was also supported by the "Advanced GPU Utilization Support Program” funded by MSIT, Korea.

\subsection*{Acknowledgement}
This work was supported by the AI Graduate School Program at POSTECH (RS-2019-II191906 (5\%)), and the NRF grants (RS-2025-24535146 (10\%), RS-2026- 25491789 (40\%)), the IITP grants (RS-2022-II220926 (25\%), RS-2026-25518317 (20\%)) funded by MSIT, Korea.
This work was also supported by the "Advanced GPU Utilization Support Program” funded by MSIT, Korea.

\bibliographystyle{splncs04}
\bibliography{main}
\section{Efficiency Analysis}
\label{sec:effi}

Our implementation is designed to minimize computational overhead during protection optimization.
For CogVideoX~\cite{cog}, we apply the proposed losses only to the middle 50\% of attention layers, which provides a favorable trade-off between computational cost and protection effectiveness.
This configuration is used for all CogVideoX experiments reported in the main paper.

Although I2VGuard~\cite{i2vguard} applies its loss to a single middle attention layer, its computational cost remains substantially higher than ours (see Table~\ref{tab:efficiency} in the main manuscript).
This is because I2VGuard requires generating a full target video and performing dual forward passes for each training sample.

For LTX-Video~\cite{ltx}, both our method and I2VGuard operate on the full set of cross-attention layers.
Nevertheless, our approach still achieves significantly lower optimization cost.

We additionally report the wall-clock time required to optimize the protection perturbation for a single image.
All experiments are measured on a single NVIDIA A100 GPU.

Our method requires approximately 7 minutes for CogVideoX and 1.5 minutes for LTX-Video, while I2VGuard requires approximately 13 minutes and 3 minutes, respectively.
These runtime measurements are consistent with the FLOPs analysis reported in the main paper and indicate that our approach substantially reduces the protection optimization cost.

\section{A Unified Perspective on the Loss Design}
\label{sec:unified}

Although our implementation differs between full-attention and cross-attention architectures, both variants are derived from the same protection objective: reducing the influence of text-conditioned signals so that visual-only pathways dominate the generation process.
The specific loss forms differ because the two architectures combine text and visual information in structurally different ways.
In this section, we show how both architectures can be interpreted through a common decomposition into text-dependent and visual-only components, and how our loss design naturally follows from this view.

\paragraph{\textbf{\textup{Full-Attention.}}}
In full-attention architectures, text-dependent and visual-only contributions are mixed within a single attention operation:
\begin{align}
z^{\ell+1} = z^{\ell} + W_o^{\ell} \left( \underbrace{\mathcal{A}_{z,z}V_z + \mathcal{A}_{z,i}V_i}_{\text{visual-only}} + \underbrace{\mathcal{A}_{z,t}V_t}_{\text{text-dependent}} \right),
\end{align}
where $z^\ell$ denotes the video latent at layer $\ell$, $W_o^\ell$ is the output projection, $\mathcal{A}_{z,\cdot}$ denotes the attention weights associated with video latent, and $V_\cdot$ is the corresponding value vector for each modality.

The key property of full attention is that these contributions compete through softmax normalization:
\begin{align}
\mathcal{A}_{z,t} = \frac{\exp(A_{z,t})}{\exp(A_{z,z}) + \exp(A_{z,i}) + \exp(A_{z,t})},
\end{align}
where $A_{z,\cdot}$ denotes the corresponding pre-softmax attention logits.
To make the text-dependent contribution $\mathcal{A}_{z,t}V_t$ small, it is natural to reduce $\mathcal{A}_{z,t}$ relative to the visual terms.
This occurs when
\begin{align}
\exp(A_{z,z}) + \exp(A_{z,i}) \gg \exp(A_{z,t}).
\end{align}

This decomposition directly motivates our loss design for full-attention models: we suppress the text-dependent score $A_{z,t}$ while simultaneously increasing the visual-only scores $A_{z,z}$ and $A_{z,i}$.
Under softmax competition, this reduces the effective contribution of text-conditioned signals and shifts attention toward visual-only pathways.

\paragraph{\textbf{\textup{Cross-Attention.}}}
In cross-attention architectures, text-dependent and visual-only operations are separated into different modules:
\begin{align}
z^{\ell+1} = z^{\ell} + \underbrace{z_{\text{SA}}^{\ell}}_{\text{visual-only}} + \underbrace{z_{\text{CA}}^{\ell}}_{\text{text-dependent}},
\end{align}
where $z_{\text{SA}}^{\ell}$ is produced by self-attention over visual tokens only, and $z_{\text{CA}}^{\ell}$ is produced by cross-attention conditioned on the text input.

Unlike full attention, the two components are combined additively rather than through softmax competition.
As a result, suppressing text influence amounts to reducing the relative contribution of the cross-attention term:
\begin{align}
\|z_{\text{CA}}^{\ell}\| \ll \|z_{\text{SA}}^{\ell}\|.
\end{align}
When the self-attention contribution dominates in magnitude, the influence of the text-conditioned branch becomes relatively small in the final representation.

This leads to the analogous design principle for cross-attention models: reduce the magnitude of the text-dependent term $\|z_{\text{CA}}^{\ell}\|$ while enlarging the visual-only term $\|z_{\text{SA}}^{\ell}\|$.
Although the mechanism differs from full attention, the objective is the same: make the final representation rely more on visual-only pathways and less on the text-conditioned branch.

\paragraph{\textbf{\textup{Unified Perspective.}}}
The two architectures, therefore, differ mainly in how text and visual information are combined.
In full attention, the competition is normalized through softmax, so relatively small changes in pre-softmax logits can be amplified into large changes in effective attention.
In cross attention, the two branches are combined linearly, so dominance is governed more directly by the relative magnitudes of the self-attention and cross-attention outputs.

Despite this structural difference, both variants of our method follow the same underlying principle: weakening text-conditioned contributions while strengthening visual-only pathways.
This unified view also helps explain the empirical trend in Table~\ref{tab:vbench} of the main paper, where full-attention architectures tend to exhibit stronger protection effects.
This behavior is consistent with the stronger competitive suppression induced by softmax-normalized attention.

\section{Evaluation Protocol}
\label{sec:evalproto}
\subsection{Complementarity with Quality Benchmarks}
\label{sec:complementarity}

We observe several mismatch cases between quality-oriented benchmarks and our Video-LLM-based failure-finding evaluation.
These examples suggest that the two evaluation paradigms capture different aspects of generated videos.

First, we observe cases where highly dynamic yet semantically plausible videos receive relatively low scores from VBench metrics~\cite{vbench,vbench++}, while the Video-LLM evaluator does not identify clear failures.
In such examples, the generated video exhibits substantial temporal variation or rapid scene changes, which can lower quality-oriented scores even when the main subject identity and scene semantics remain largely consistent.

Conversely, we also observe cases where VBench metrics remain relatively favorable even though the generated video exhibits semantic inconsistencies, structural distortions, or physically implausible motion over time.
This is plausible because VBench primarily measures perceptual quality and temporal continuity, whereas a Video-LLM is trained for video understanding and can better capture higher-level failure signals related to object identity, structure, action, and physical plausibility.

To illustrate these complementary behaviors, we provide representative mismatch examples on the project page.
The visualization contains twelve cases, including both VBench-low/Qwen-high and VBench-high/Qwen-low examples.

For ease of interpretation, each video example reports both raw and normalized scores.
The normalized score for each dimension is computed by percentile-based ranking over the full pool of generated videos used in our experiments, including both protected and unprotected samples across all evaluated settings.
This allows each example to be interpreted relative to the broader evaluation distribution rather than only within the selected mismatch subset.

Each video example contains the following information.
The left panel shows the video title (prompt), the average Qwen score (orange bar), and short textual descriptions from the Video-LLM evaluator.
For Qwen-low cases, these descriptions highlight the failure dimensions identified by the evaluator, whereas for Qwen-high cases they provide concise summaries of the generated video content.
The center panel shows the generated video itself.
The right panel shows the normalized average VBench score and the three most influential VBench dimensions for that sample, including both raw metric values and percentile-normalized scores.

\subsection{Human Study Alignment}
\label{sec:humanstudy}
Table~\ref{tab:human_study} of the main paper reports the human preference results and their alignment with our evaluation protocol.
Here, we provide additional details on the study design, including participant recruitment, triplet construction, blinding, randomization, and the mapping between our protocol dimensions and human evaluation criteria.

We conduct a user study to verify that the proposed evaluation protocol aligns with human judgment.
Thirty participants are recruited from the computer vision community, and each participant evaluates 20 video triplets sampled from the full set of 355 generations.
Each triplet contains three videos generated from the same input image and text prompt, corresponding to \emph{Clean}, \emph{I2VGuard}, and \emph{Ours}.
The study is fully blinded: method identities are hidden from participants, and the presentation order of the three videos is independently randomized for each triplet to mitigate position bias.
For each triplet, participants select the \emph{best} and \emph{worst} results in terms of \emph{consistency} and \emph{plausibility}, following the provided instructions.

To reduce the cognitive burden of human evaluation, we do not ask participants to separately assess all four dimensions used in our protocol.
Instead, we group them into two higher-level concepts that are more natural for human comparison.
Specifically, \emph{consistency} reflects whether the main subject and its structure remain faithful and stable throughout the video, and therefore primarily corresponds to \emph{subject preservation} and \emph{structural consistency}.
In contrast, \emph{plausibility} reflects whether the generated video appears temporally natural and visually coherent overall, and thus mainly relates to \emph{dynamic consistency} and \emph{artifact suppression}.
This simplification is used only for the human study to obtain reliable comparative judgments across multiple videos, while our full protocol retains four interpretable dimensions for finer-grained failure analysis.

Due to the cognitive load of video comparison, each participant evaluates a representative subset of 20 triplets, focusing on dynamic scenes and human activities.
Responses marked as \emph{No preference} are excluded from the preference statistics.

\begin{table}[t]
\centering
\footnotesize
\caption{Stability of the Video-LLM evaluator~\cite{qwen3vl} across repeated runs.
We report the mean and standard deviation over five independent evaluations of the same videos generated from protected images.
All evaluations use stochastic decoding with temperature 0.2, top-$p$ 0.9, and top-$k$ 20.}
\label{tab:stability}
\begin{tabular}{
l
S[table-format=1.3]
@{\hspace{2pt}{\color{gray}\scriptsize$\pm$}\hspace{2pt}}
S[table-format=1.4]
S[table-format=1.3]
@{\hspace{2pt}{\color{gray}\scriptsize$\pm$}\hspace{2pt}}
S[table-format=1.4]
}
\toprule
& \multicolumn{2}{c}{CogVideoX} & \multicolumn{2}{c}{LTX-Video} \\
\cmidrule(lr){2-3}\cmidrule(lr){4-5}
Dimension & \multicolumn{2}{c}{Mean $\pm$ Std.} & \multicolumn{2}{c}{Mean $\pm$ Std.} \\
\midrule
\rowcolor{gray!8}
\textbf{Overall}
& 3.251 & 0.0161
& 4.380 & 0.0091 \\
\midrule
Subject Preservation
& 4.366 & 0.0380
& 4.056 & 0.0321 \\
Structural Consistency
& 3.239 & 0.0491
& 3.017 & 0.0136 \\
Dynamic Consistency
& 2.907 & 0.0195
& 2.845 & 0.0268 \\
Artifact Suppression
& 2.494 & 0.0126
& 2.580 & 0.0100 \\
\bottomrule
\end{tabular}
\end{table}

\begin{table}[t]
\centering
\scriptsize
\caption{Stability and agreement of Video-LLM evaluator~\cite{qwen3vl} across repeated stochastic runs. 
We report multi-run agreement using Kendall's $W$~\cite{kendallW} and average pairwise correlations using Pearson's $r$~\cite{pearson}, Spearman's $\rho$~\cite{spearman}, and Kendall's $\tau_b$~\cite{kendall}, computed over five independent evaluations.}
\label{tab:agreement}
\begin{tabular}{l S[table-format=1.6] S[table-format=1.6] S[table-format=1.6] S[table-format=1.6]
                S[table-format=1.6] S[table-format=1.6] S[table-format=1.6] S[table-format=1.6]}
\toprule
& \multicolumn{4}{c}{CogVideoX} 
& \multicolumn{4}{c}{LTX-Video} \\
\cmidrule(lr){2-5}\cmidrule(lr){6-9}
Dimension 
& {$W$} & {$r$} & {$\rho$} & {$\tau_b$}
& {$W$} & {$r$} & {$\rho$} & {$\tau_b$} \\
\midrule
\rowcolor{gray!8}
\textbf{Overall} 
& 0.9379 & 0.9299 & 0.9223 & 0.8220
& 0.9695 & 0.9646 & 0.9618 & 0.8827 \\
\midrule
Subject Preservation
& 0.8511 & 0.8250 & 0.8141 & 0.7840
& 0.9158 & 0.9151 & 0.8947 & 0.8556 \\
Structural Consistency
& 0.8807 & 0.8561 & 0.8508 & 0.8009
& 0.9231 & 0.9210 & 0.9038 & 0.8564 \\
Dynamic Consistency
& 0.8057 & 0.7997 & 0.7571 & 0.7042
& 0.8539 & 0.8690 & 0.8175 & 0.7669 \\
Artifact Suppression
& 0.8561 & 0.8788 & 0.8201 & 0.7859
& 0.8803 & 0.9032 & 0.8504 & 0.8079 \\
\bottomrule
\end{tabular}
\end{table}

\subsection{Evaluator Consistency Analysis}
\label{sec:repeat}

While Section~\ref{sec:eval_proto} introduces the failure-finding evaluation protocol, here we
examine its empirical stability.
Unlike conventional video quality grading, which often requires subtle perceptual judgments, the proposed protocol focuses on detecting clear failure evidence grounded in observable frames.
This formulation makes the evaluation less sensitive to small variations in the evaluator's output and reduces reliance on fine-grained subjective interpretation.

An additional practical advantage of our protocol is that it uses a unified evaluation query and rubric across all samples.
Unlike QA-style benchmarks~\cite{vbench2, vidbench} that often rely on sample-specific questions, our formulation evaluates every generated video using the same failure criteria, making the protocol easier to apply beyond a fixed benchmark.

Empirically, we observe that the evaluation outcomes remain highly stable across repeated runs.
When running the same Video-LLM evaluator~\cite{qwen3vl} multiple times on identical videos, the resulting scores exhibit very low variance, with standard deviations below 0.05 on a 5-point scale across all dimensions in Table~\ref{tab:stability}.
These results suggest that the proposed failure-finding formulation provides stable evaluation trends even under stochastic decoding.

In addition to score variance, we also examine whether the relative ranking of methods remains consistent across repeated runs.
Despite stochastic decoding, the evaluator produces generally consistent method orderings across independent evaluations (Table~\ref{tab:agreement}), suggesting that the proposed protocol yields stable comparative trends.

\begin{table}[t]
\centering
\footnotesize
\caption{Cross-evaluator consistency measured using rank-based metrics.
We compare the rankings induced by different Video-LLM evaluators on the same set of generated videos.
High correlation and pairwise agreement indicate that the relative ordering of methods remains consistent across evaluators.}
\label{tab:cross_eval_consistency}
\begin{tabular}{lccc}
\toprule
Evaluator Pair & Spearman $\rho$ & Kendall $\tau_b$ & Pairwise Agreement \\
\midrule
Qwen -- GPT        & 0.965 & 0.869 & 0.935 \\
Qwen -- VideoLLaMA3 & 0.920 & 0.791 & 0.895 \\
Qwen -- InternVL3  & 0.950 & 0.843 & 0.922 \\
GPT -- VideoLLaMA3  & 0.965 & 0.869 & 0.935 \\
GPT -- InternVL3   & 0.977 & 0.895 & 0.948 \\
VideoLLaMA3 -- InternVL3 & 0.969 & 0.895 & 0.948 \\
\midrule
\textbf{Multiway Kendall's $W$} & \multicolumn{3}{c}{0.968} \\
\bottomrule
\end{tabular}
\end{table}

\begin{table}[t]
\centering
\footnotesize
\caption{Dimension-wise multiway agreement across evaluators measured by Kendall's coefficient of concordance ($W$).
High values indicate that different evaluators induce similar method rankings for each failure dimension.
}
\label{tab:dimension_agreement}
\begin{tabular}{lc}
\toprule
Dimension & Kendall's $W$ \\
\midrule
Subject Preservation & 0.953 \\
Structural Consistency & 0.972 \\
Dynamic Consistency & 0.919 \\
Artifact Suppression & 0.928 \\
\bottomrule
\end{tabular}
\end{table}

\subsection{Cross-Evaluator Consistency}
\label{sec:crosseval}

We use Qwen3-VL-8B~\cite{qwen3vl} as the primary evaluator.
To assess robustness to evaluator choice, we additionally include VideoLLaMA3-7B~\cite{videollama3} and InternVL3-9B~\cite{intern} as open-source evaluators, and GPT-5-mini~\cite{gpt5} as a commercial evaluator for cross-model validation.

The evaluation set is constructed from all 355 source images in the VBench Image-to-Video dataset.
For each source image, we collect 20 generated videos covering clean generation as well as white-box protection, black-box protection, and purification-based pipelines for both \emph{Ours} and \emph{I2VGuard}.

All evaluators are applied to the same set of generated videos using the same evaluation prompt and rubric.
As shown in Table~\ref{tab:cross_eval_consistency}, the evaluator-induced rankings exhibit high agreement across models.
We also observe consistently high multiway agreement across evaluators and across individual dimensions (Table~\ref{tab:dimension_agreement}).
These results suggest that the proposed failure-finding protocol yields stable comparative conclusions across different Video-LLM evaluators.

\label{sec:promptfree}
\begin{table*}[]
\scriptsize
\centering
\caption{Additional experiments on the absence of text prompts. $\ast$ denotes perturbations optimized without text prompts.}
\resizebox{\linewidth}{!}{
\begin{tabular}{l cccccccc}
\toprule
{Method} &
\makecell{I2V\\Subject} &
\makecell{I2V\\Background} &
\makecell{Subject\\Consistency} &
\makecell{Background\\Consistency} &
\makecell{Aesthetic\\Quality} &
\makecell{Imaging\\Quality} &
\makecell{Temporal\\Flickering} &
\makecell{Motion\\Smoothness} \\
\midrule
{\textcolor{gray}{\textit{CogVideoX~\cite{cog}}}} & & & & & & & & \\
I2VGuard~\cite{i2vguard} & 94.58 & 97.14 & 93.32 & 96.17 & 58.61 & 65.30 & 96.47 & 97.61 \\
Ours$^\ast$ & 92.37 & 95.67 & 89.56 & 94.40 & 56.08 & 66.51 & 94.96 & 96.19 \\
Ours & \textbf{91.20} & \textbf{94.28} & \textbf{87.54} & \textbf{93.16} & \textbf{54.72} & \textbf{63.71} & \textbf{94.47} & \textbf{95.66} \\
\hline
{\textcolor{gray}{\textit{LTX-Video~\cite{ltx}}}} & & & & & & & & \\
I2VGuard~\cite{i2vguard}  & 94.36 & 95.56 & 90.25 & 95.00 & 56.28 & 62.63 & 98.23 & 98.90 \\
Ours$^\ast$  & 93.76 & 95.28 & 89.52 & 94.78 & 55.01 & 61.66 & 97.98 & 98.71 \\
Ours  & \textbf{93.64} & \textbf{95.07} & \textbf{89.36} & \textbf{94.99} & \textbf{54.99} & \textbf{61.44} & \textbf{97.95} & \textbf{98.67} \\
\bottomrule
\end{tabular}
}
\label{tab:prompt_ablation}
\end{table*}

\begin{table}[t]
\scriptsize
\setlength{\tabcolsep}{4pt}
\centering
\caption{Additional experiments on the absence of text prompts with our evaluation protocol. $\ast$ denotes perturbations optimized without text prompts.}
\resizebox{0.9\linewidth}{!}{
\begin{tabular}{@{}lccccc@{}}
\toprule
Method
& \makecell{Subject\\Preservation}
& \makecell{Structural\\Consistency}
& \makecell{Dynamic\\Consistency}
& \makecell{Artifact\\Suppression}
& \makecell{Average} \\
\midrule
\multicolumn{6}{@{}l}{\textcolor{gray}{\textit{CogVideoX~\cite{cog}}}}\\
I2VGuard & 4.72 & 4.18 & 3.81 & 2.97 & 3.92 \\
Ours$^\ast$ & 4.57 & 3.49 & 3.18 & 2.84 & 3.52 \\
Ours & \textbf{4.37} & \textbf{3.24} & \textbf{2.91} & \textbf{2.49} & \textbf{3.25} \\
\midrule
\multicolumn{6}{@{}l}{\textcolor{gray}{\textit{LTX-Video~\cite{ltx}}}}\\
I2VGuard & 4.24 & 3.18 & 2.99 & 2.70 & 3.28 \\
Ours$^\ast$ & 4.16 & 3.14 & 2.96 & 2.66 & 3.23 \\
Ours & \textbf{4.06} & \textbf{3.02} & \textbf{2.85} & \textbf{2.58} & \textbf{3.12} \\
\bottomrule
\end{tabular}
}
\label{tab:prompt_ours_proto}
\end{table}

\section{Additional Experiments}
\label{sec:addex}
\subsection{Robustness under Prompt-Free Settings}

Given that our method aims to suppress text influence, an important question is whether the protection effect depends on the specific prompts used during perturbation generation.
To investigate this, we conduct an ablation study in which perturbations are optimized using empty prompts, representing an extreme prompt-agnostic setting.
We compare this with prior work and our full method with prompts in Table~\ref{tab:prompt_ablation}, and further report the results under our failure-finding evaluation protocol in Table~\ref{tab:prompt_ours_proto}.

Even without textual input during optimization, our method remains more effective than the baseline across both VBench metrics and our evaluation protocol.
This suggests that the proposed objective does not rely heavily on the specific prompts used during perturbation generation.
Instead, it appears to target architectural pathways through which text affects generation, allowing the protection effect to generalize beyond the prompts seen during optimization.

\label{sec:purification}
\begin{table}[]
\scriptsize
\centering
\caption{Robustness to purification defenses measured by VBench metrics.
We consider Crop-and-Resize (retain ratio: 80\% in both width and height),
JPEG compression (quality factor: 60), and ADVClean~\cite{advclean} before
video generation. Lower scores indicate stronger protection.}
\resizebox{\linewidth}{!}{
\begin{tabular}{l cccccccc}
\toprule
{Method} &
\makecell{I2V\\Subject} &
\makecell{I2V\\Background} &
\makecell{Subject\\Consistency} &
\makecell{Background\\Consistency} &
\makecell{Aesthetic\\Quality} &
\makecell{Imaging\\Quality} &
\makecell{Temporal\\Flickering} &
\makecell{Motion\\Smoothness}\\
\midrule
\multicolumn{9}{@{}l}{\textcolor{gray}{\textit{Crop and Resize~\cite{crop} + CogVideoX~\cite{cog}}}}\\
I2VGuard~\cite{i2vguard} & 92.92 & 94.95 & 94.15 & 96.46 &59.51 & \textbf{65.60} & 96.40 & 97.61 \\
Ours & \textbf{92.89} & \textbf{94.89} & \textbf{93.46} & \textbf{96.30} & \textbf{58.26} & 67.53 & \textbf{95.95} & \textbf{97.22}\\
\cdashline{1-9}[1pt/1pt]
\multicolumn{9}{@{}l}{\textcolor{gray}{\textit{JPEG Compression~\cite{crop} + CogVideoX~\cite{cog}}}}\\
I2VGuard~\cite{i2vguard} & 93.95 & 96.44 & 92.92 & 95.66 & 58.78 & \textbf{65.38} & 96.03 & 97.25\\
Ours & \textbf{93.85} & \textbf{96.30} & \textbf{91.96} & \textbf{95.15} & \textbf{57.45} & 67.10 & \textbf{95.48} & \textbf{96.78}\\
\cdashline{1-9}[1pt/1pt]
\multicolumn{9}{@{}l}{\textcolor{gray}{\textit{ADVClean~\cite{advclean} + CogVideoX~\cite{cog}}}}\\
I2VGuard~\cite{i2vguard} & 94.94 & 97.48 & 93.79 & 96.27 & \textbf{56.97} & \textbf{66.02} & 96.75 & 97.69\\
Ours & \textbf{93.71} & \textbf{97.07} & \textbf{91.84} & \textbf{95.71} & \textbf{56.97} & 67.04 & \textbf{95.65} & \textbf{96.83}\\
\hline
\multicolumn{9}{@{}l}{\textcolor{gray}{\textit{Crop and Resize~\cite{crop} + LTX-Video~\cite{ltx}}}}\\
I2VGuard~\cite{i2vguard} & 93.32 & 94.32 & 90.36 & 95.11 & 55.68 & 61.82 & 98.21 & 98.90 \\
Ours & \textbf{92.19} & \textbf{93.65} & \textbf{88.59} & \textbf{94.71} & \textbf{53.92} & \textbf{59.96} & \textbf{97.79} & \textbf{98.56}\\
\cdashline{1-9}[1pt/1pt]
\multicolumn{9}{@{}l}{\textcolor{gray}{\textit{JPEG Compression~\cite{crop} + LTX-Video~\cite{ltx}}}}\\
I2VGuard~\cite{i2vguard} & 92.04 & 94.60 & 85.71 & 93.48 & 53.03 & 58.69 & 97.38 & 98.14 \\
Ours & \textbf{91.58} & \textbf{93.89} & \textbf{85.46} & \textbf{93.23} & \textbf{52.02} & \textbf{57.96} & \textbf{97.25} & \textbf{98.02}\\
\cdashline{1-9}[1pt/1pt]
\multicolumn{9}{@{}l}{\textcolor{gray}{\textit{ADVClean~\cite{advclean} + LTX-Video~\cite{ltx}}}}\\
I2VGuard~\cite{i2vguard} & 96.08 & 96.71 & 93.71 & 96.92 & 58.59 & 64.31 & 98.97 & 99.33\\
Ours & \textbf{95.78} & \textbf{96.44} & \textbf{92.94} & \textbf{96.67} & \textbf{57.31} & \textbf{63.35} & \textbf{98.76} & \textbf{99.20}\\
\bottomrule
\end{tabular}
}
\label{tab:purification}
\end{table}
\begin{table}[t]
\scriptsize
\setlength{\tabcolsep}{4pt}
\centering
\caption{Robustness to purification defenses measured by our failure-finding
evaluation protocol.}
\resizebox{0.8\linewidth}{!}{
\begin{tabular}{@{}lccccc@{}}
\toprule
Method
& \makecell{Subject\\Preservation}
& \makecell{Structural\\Consistency}
& \makecell{Dynamic\\Consistency}
& \makecell{Artifact\\Suppression}
& \makecell{Average} \\
\midrule
\multicolumn{6}{@{}l}{\textcolor{gray}{\textit{Crop and Resize~\cite{crop} + CogVideoX~\cite{cog}}}}\\
I2VGuard & 4.78 & 4.26 & 3.93 & \textbf{3.27} & 4.06 \\
Ours & \textbf{4.75} & \textbf{4.20} & \textbf{3.85} & 3.41 & \textbf{4.05} \\
\cdashline{1-6}[1pt/1pt]
\multicolumn{6}{@{}l}{\textcolor{gray}{\textit{JPEG Compression~\cite{crop} + CogVideoX~\cite{cog}}}}\\
I2VGuard & \textbf{4.71} & 4.16 & 3.78 & \textbf{3.44} & 4.03 \\
Ours & \textbf{4.71} & \textbf{4.02} & \textbf{3.70} & 3.59 & \textbf{4.01} \\
\cdashline{1-6}[1pt/1pt]
\multicolumn{6}{@{}l}{\textcolor{gray}{\textit{ADVClean~\cite{advclean} + CogVideoX~\cite{cog}}}}\\
I2VGuard & 4.72 & 4.18 & 3.86 & \textbf{3.94} & 4.17 \\
Ours & \textbf{4.61} & \textbf{3.79} & \textbf{3.65} & 3.96 & \textbf{4.00} \\
\midrule
\multicolumn{6}{@{}l}{\textcolor{gray}{\textit{Crop and Resize~\cite{crop} + LTX-Video~\cite{ltx}}}}\\
I2VGuard & 4.23 & 3.23 & 3.03 & 2.60 & 3.27 \\
Ours & \textbf{4.01} & \textbf{3.01} & \textbf{2.85} & \textbf{2.44} & \textbf{3.08} \\
\cdashline{1-6}[1pt/1pt]
\multicolumn{6}{@{}l}{\textcolor{gray}{\textit{JPEG Compression~\cite{crop} + LTX-Video~\cite{ltx}}}}\\
I2VGuard & 3.65 & 2.72 & 2.65 & 2.51 & 2.88 \\
Ours & \textbf{3.62} & \textbf{2.60} & \textbf{2.51} & \textbf{2.47} & \textbf{2.80} \\
\cdashline{1-6}[1pt/1pt]
\multicolumn{6}{@{}l}{\textcolor{gray}{\textit{ADVClean~\cite{advclean} + LTX-Video~\cite{ltx}}}}\\
I2VGuard & 4.46 & 3.62 & 3.32 & 3.62 & 3.76 \\
Ours & \textbf{4.31} & \textbf{3.44} & \textbf{3.28} & \textbf{3.43} & \textbf{3.62} \\
\bottomrule
\end{tabular}
}
\label{tab:purification_ours_proto}
\end{table}

\subsection{Robustness to Purification}

To evaluate robustness under purification defenses, we apply post-processing techniques, including Crop-and-Resize~\cite{crop}, JPEG compression~\cite{crop}, and ADVClean~\cite{advclean} to protected images before video generation.
We report the comparison under purification using both VBench metrics (Table~\ref{tab:purification}) and our failure-finding evaluation protocol (Table~\ref{tab:purification_ours_proto}).

We note that the purification settings used here are relatively aggressive. In particular, Crop-and-Resize keeps only 80\% of the original width and height before resizing back, JPEG compression is applied with a quality factor of 60, and ADVClean also constitutes a strong purification setting.
Accordingly, the absolute score drop under purification should be interpreted with this severity in mind.

Despite the overall degradation caused by purification, our method remains more effective than I2VGuard on most semantic, structural, and motion-related dimensions.
At the same time, we observe that I2VGuard sometimes attains lower scores on artifact-sensitive dimensions such as Imaging Quality and Artifact Suppression, especially for CogVideoX.

A plausible explanation is that, for CogVideoX, I2VGuard starts from more perceptible perturbations, which is also consistent with the visibility metrics reported in Table~\ref{tab:imper} of the main paper (higher DISTS/LPIPS and lower PSNR/SSIM for CogVideoX).
As a result, purification may still leave stronger residual high-frequency patterns for I2VGuard, which can be penalized heavily by dimensions such as Imaging Quality and Artifact Suppression.

\begin{figure}[t]
\centering
    \includegraphics[width=0.8\linewidth]{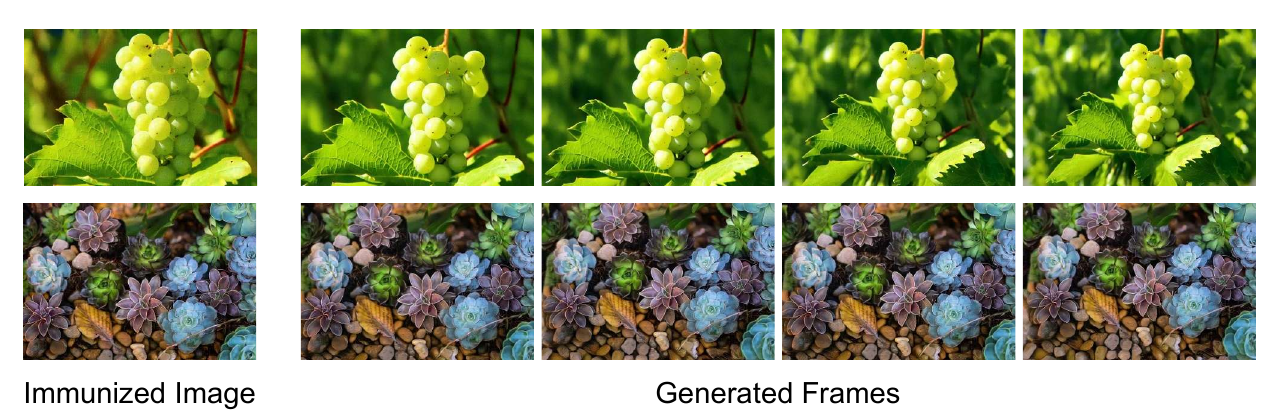}
    \caption{Failure cases on static or low-motion scenes. Our method shows reduced effectiveness when the expected video dynamics are minimal, making visible animation failures harder to induce.}
    \label{fig:failure}
    
\end{figure}

\begin{figure*}[t]
\centering
    \includegraphics[width=\linewidth]{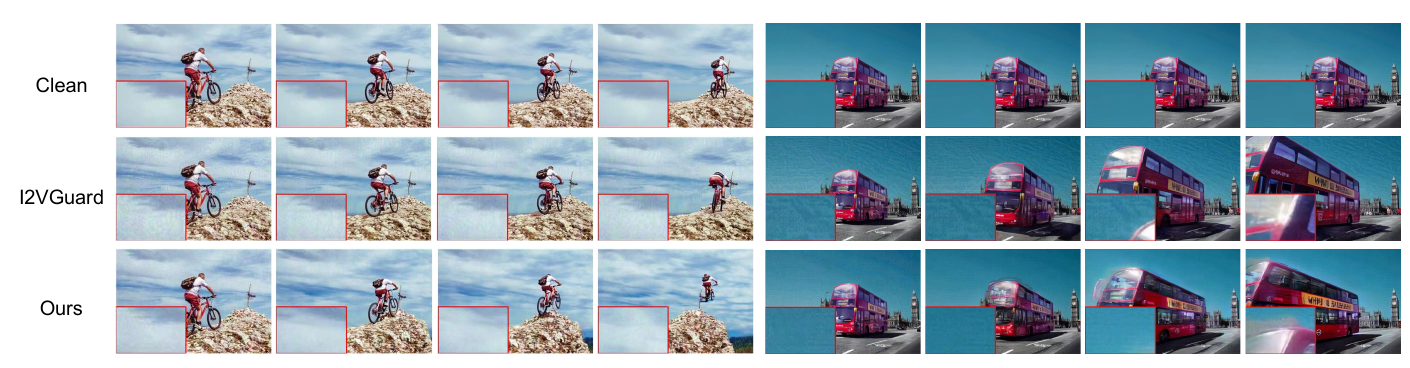}
    \caption{Generation results from JPEG-purified images using CogVideoX. After purification, I2VGuard often leaves residual high-frequency noise, whereas our method more often leads to structural collapse or abnormal motion. This illustrates that purification can weaken both methods while altering their dominant failure modes in different ways.}
    \label{fig:purification_qual}
    
\end{figure*}

\begin{figure*}[t]
\centering
    \includegraphics[width=\linewidth]{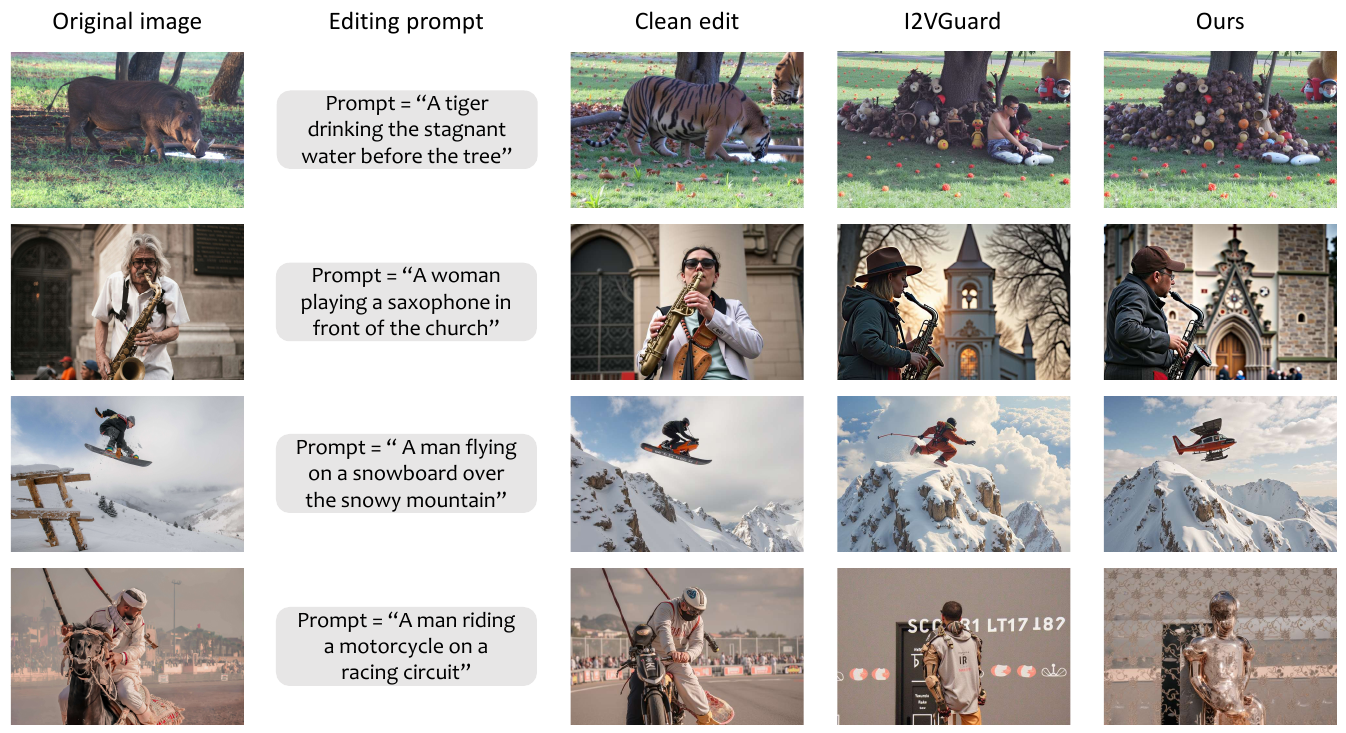}
\caption{Qualitative examples on an image editing model (FLUX.1~\cite{flux}).
Although not systematically evaluated, we observe that protected images can lead to editing results that differ from clean editing results.
}
    \label{fig:editing}
    
\end{figure*}

\section{Limitation}
\label{sec:limit}

Our method has several limitations.
First, although it is more efficient than I2VGuard, the proposed protection still relies on per-image optimization and therefore cannot provide instantaneous protection.
Developing universal protection mechanisms~\cite{uap, uapimm} remains an important direction for future work.

Second, the protection effect can be reduced in static or low-motion scenes.
When the prompt induces minimal motion, the generated video may remain visually close to the input image, making it more difficult for protection signals to manifest as clear animation failures (Figure~\ref{fig:failure}).
This limitation is partly inherent to failure-based protection, since visible disruption is harder to expose when the desired animation itself is minimal.

Third, purification techniques such as JPEG compression or adversarial cleaning can partially weaken the protection effect (Table~\ref{tab:purification}).
This highlights the difficulty of maintaining protection effectiveness under post-processing defenses.
As shown in Figure~\ref{fig:purification_qual}, purification can also change the dominant failure mode: I2VGuard often leaves residual high-frequency noise, whereas our method more often induces structural collapse or abnormal motion.
Recent image protection methods for editing and inpainting have explored purification-aware designs to improve robustness against post-processing defenses, and adapting similar ideas to text-guided image-to-video protection may be a promising direction for future work.

\section{Discussion}
\label{sec:disc}

Our evaluation adopts a failure-finding perspective rather than a traditional video quality benchmarking framework.
Instead of measuring fine-grained perceptual quality, the goal is to identify clear signals that image-faithful animation has broken down.

Interpreting artifacts, therefore, requires some nuance.
Mild artifacts, such as residual noise introduced by purification, may degrade perceptual quality without necessarily indicating a semantic failure of image-faithful animation.
In contrast, severe artifacts that dominate the frame, occlude the subject, or disrupt scene structure typically correspond to clear generation breakdowns.
Because failure can manifest in multiple ways, a single binary criterion does not always capture its full range.

Within this framework, the average of dimension scores serves as a practical summary
of failure severity.
Such averages can still reflect protection effectiveness reasonably well, but they should not be interpreted as a definitive definition of failure.
Rather, they provide an operational summary of failure trends, while the broader question of what constitutes a clear failure in image-to-video generation remains open.

As an additional qualitative result, we examine protected images with an image editing model (FLUX.1~\cite{flux}).
As shown in Figure~\ref{fig:editing}, protected inputs can produce editing results that differ noticeably from the corresponding clean edits.
Since this effect is not systematically evaluated in this work, we leave a more careful investigation to future work.

\section{Broader Impact}

This work is motivated by the growing risk of unintended reuse of publicly shared images in text-guided image-to-video generation systems.
By weakening prompt-conditioned generation from a given image, our approach aims to help content owners better protect their images from producing misleading or undesired videos.

However, it operates by protecting the image itself rather than filtering prompts.
As a result, it may also weaken benign or authorized prompt-conditioned uses of the same image.
Developing more selective protection mechanisms that better distinguish malicious use from benign use remains an important direction for future work.

\begin{algorithm}[!t]
\caption{Image Protection against Text-Guided I2V}
\label{alg:i2v_attack}
\textbf{Input:} Image $x$, uninformative target image $x_{tar}$, text condition $c_{\text{txt}}$ (optional; if absent, use $\emptyset$), generative model $\mathcal{G}$, VAE encoder $\mathcal{E}$, step size $\alpha$, perturbation bound $\epsilon$, number of outer iterations $N$, number of denoising steps $T$, total number of PGD updates $K$ with $K = N \times T$. \\
\textbf{Output:} Protected image $x_{adv}$.
\begin{algorithmic}[1]
    \STATE Initialize perturbation $\delta_0 \sim \mathcal{U}(-\epsilon, \epsilon)$
    \STATE Initialize update counter $k = 0$
    \STATE Precompute target latent $z_{tar} = \mathcal{E}(x_{tar})$
    \FOR{$n = 1$ to $N$}
        \STATE Initialize video latent $z_T \sim \mathcal{N}(0,I)$
        \FOR{$t = T$ to $1$}
            \STATE $x_{adv} = \mathrm{Clip}(x + \delta_k, 0, 1)$
            \STATE $z_{adv} = \mathcal{E}(x_{adv})$
            \STATE $\epsilon_\theta = \mathcal{G}(z_t, z_{adv}, c_{\text{txt}}, t)$
            \STATE Compute total loss $\mathcal{L}(z_{adv}, z_{tar})$ via Eq.~(\textcolor{red}{10})
            \STATE $g_k = \nabla_{\delta_k}\mathcal{L}$
            \STATE $\delta_{k+1} = \delta_k - \alpha \cdot \mathrm{sign}(g_k)$
            \STATE $\delta_{k+1} = \mathrm{Clip}(\delta_{k+1}, -\epsilon, \epsilon)$
            \STATE $z_{t-1} = \mathrm{SchedulerStep}(z_t, \epsilon_\theta, t)$
            \STATE $k = k + 1$
        \ENDFOR
    \ENDFOR
    \STATE $x_{adv} = \mathrm{Clip}(x + \delta_K, 0, 1)$
    \STATE \textbf{return} $x_{adv}$
\end{algorithmic}
\end{algorithm}

\section{Reproducibility}
\label{sec:reproduce}

To facilitate reproducibility, we provide the key components required to reproduce our experiments.
Algorithm~\ref{alg:i2v_attack} describes the protection optimization procedure used to generate protected images.
Table~\ref{tab:unseen_prompt_examples} presents representative examples of unseen prompt construction used in robustness evaluation.
Figures~\ref{fig:appendix_prompt_template}--\ref{fig:appendix_artifact_suppression} show the shared Video-LLM evaluation prompt template together with the dimension-specific scoring criteria.

The complete source code and evaluation pipeline are provided in the supplementary material package included with this submission.
Hyperparameters for protection optimization and video generation are reported in the main paper.
\begin{table}[t]
\centering
\scriptsize
\setlength{\tabcolsep}{4pt}
\caption{Representative examples of unseen prompt construction.
Each unseen prompt introduces a new motion instruction while preserving the original scene semantics.}
\resizebox{0.9\linewidth}{!}{
\begin{tabular}{c >{\raggedright\arraybackslash}p{5.0cm} >{\raggedright\arraybackslash}p{6.0cm}}
\toprule
\textbf{\#} & \textbf{Original Prompt} & \textbf{Unseen Prompt} \\
\midrule
1 & A little girl sits quietly on a bus
  & A little girl makes a quick gesture, then relaxes \\

2 & A man rides a bike down a street
  & A man rides forward, then briefly turns around \\

3 & A woman carries plants over her head
  & A woman speeds up, then slows and shifts direction \\

4 & A tiger walks through a wooded area
  & A tiger trots ahead, then suddenly changes direction \\

5 & A giraffe walks in a field
  & A giraffe bounds ahead and briefly shakes its body \\

6 & A small monkey holds food in its mouth
  & A small monkey darts forward, then pauses and turns its head \\

7 & A blue train travels through a green area
  & A blue train accelerates, then eases to a steady pace \\

8 & A red sports car drives through sand
  & A red sports car speeds up, then slows and shifts lanes \\
\bottomrule
\end{tabular}
}
\label{tab:unseen_prompt_examples}
\end{table}
\clearpage
\begin{figure}[t]
\centering
\fcolorbox{black}{gray!10}{%
\begin{minipage}{0.95\linewidth}
\scriptsize
\textbf{Video-LLM Evaluation Prompt Template}

\textbf{System Message:}

You are an expert evaluator for AI-generated Image-to-Video outputs.

\textbf{Strict Protocol:}
\begin{enumerate}[leftmargin=*,nosep]
    \item First, provide up to three concrete observations with timestamps or frame indices.
    \item Then, and only then, assign a discrete score from 1 to 5 based only on those observations.
    \item Assign the lowest score supported by the observed evidence.
    \item If no clear failure is observed, assign 5.
    \item If evidence is insufficient or ambiguous, set \texttt{"uncertain": true} and use score 3.
    \item Output only valid JSON.
\end{enumerate}

Do not speculate beyond what is visible.

\textbf{Input Context Provided to the Model:}
\begin{itemize}[leftmargin=*,nosep]
    \item \texttt{Video ID: \$VIDEO\_ID}
    \item \texttt{Prompt (context only): \$PROMPT\_STR}
    \item \texttt{Assume FIRST frame is the input image.}
    \item A sequence of sampled frames, each preceded by a textual marker such as
    \texttt{Frame \$FRAME\_INDEX (t=\$TIMESTAMP s)}
\end{itemize}

\textbf{Dimension-Specific Fields Injected into the Shared Template:}
\begin{itemize}[leftmargin=*,nosep]
    \item \texttt{Dimension: \$DIMENSION\_NAME}
    \item \texttt{Definition: \$DIMENSION\_DEFINITION}
    \item \texttt{Anti-contamination rule: \$ANTI\_CONTAMINATION\_RULE}
    \item \texttt{Representative failure signs: \$FAILURE\_SIGNS}
    \item \texttt{Scoring rubric (1--5, discrete): \$RUBRIC\_TEXT}
\end{itemize}

\textbf{Task Instruction:}
\begin{enumerate}[leftmargin=*,nosep]
    \item Provide up to three concrete observations with timestamps or frame indices.
    \item Based only on those observations, assign the score.
    \item If uncertain due to insufficient evidence, set \texttt{"uncertain": true} and use score 3.
\end{enumerate}

\textbf{Required Output Schema:}

{\ttfamily
\begin{tabular}{l}
\{ \\
\quad "observations": [ \\
\quad\quad \{"time": "e.g., 2.0s or frame\#25", \\
\quad\quad\quad "description": "what is visibly observed"\} \\
\quad ], \\
\quad "score": 1, \\
\quad "uncertain": false \\
\}
\end{tabular}
}

\end{minipage}
}
\caption{Shared prompt template used for Video-LLM evaluator.}
\label{fig:appendix_prompt_template}
\end{figure}

\noindent\begin{minipage}{\linewidth}
\centering
\fcolorbox{black}{gray!10}{%
\begin{minipage}{0.95\linewidth}
\scriptsize
\textbf{Subject Preservation}

\textbf{Definition:} Identity-level consistency: whether the main subject from the first frame remains present and recognizable.

\textbf{Anti-contamination rule:} Evaluate only subject identity and presence. Ignore rendering noise, blur, flicker, motion smoothness, boundary tearing, and geometry deformation unless they directly make the subject unrecognizable or replaced.

\textbf{Representative failure signs:}
\begin{itemize}[leftmargin=*,nosep]
    \item Main subject disappears
    \item Subject is replaced by a different object, person, or class
    \item Major identity drift in face, clothing, color, or material
    \item Unexpected duplication or splitting of the subject
    \item Subject merges into the background or becomes unrecognizable
\end{itemize}

\textbf{Rubric:}
\begin{itemize}[leftmargin=*,nosep]
    \item 5: Subject identity fully preserved across sampled frames
    \item 4: Minor identity drift, but still clearly the same subject
    \item 3: Noticeable identity inconsistencies affecting recognition
    \item 2: Major identity change or partial disappearance
    \item 1: Subject missing or replaced
\end{itemize}
\end{minipage}
}
\captionof{figure}{\textit{Subject Preservation} evaluation criteria.}
\label{fig:appendix_subject_preservation}
\end{minipage}

\noindent\begin{minipage}{\linewidth}
\centering
\fcolorbox{black}{gray!10}{%
\begin{minipage}{0.95\linewidth}
\scriptsize
\textbf{Structural Consistency}

\textbf{Definition:} Geometry- and boundary-level stability: whether the subject's shape, proportions, and boundaries remain consistent across frames.

\textbf{Anti-contamination rule:} Evaluate only shape, geometry, and boundary stability. Ignore global flicker or brightness shifts, generic blur or noise, and physical plausibility unless they manifest as structural deformation.

\textbf{Representative failure signs:}
\begin{itemize}[leftmargin=*,nosep]
    \item Implausible stretching or shrinking of proportions
    \item Parts appear or disappear, such as extra limbs or missing parts
    \item Holes, tearing, or fragmentation in the silhouette
    \item Boundary tearing, ghosting, or halo trails
    \item Rigid objects wobble, melt, or undergo severe geometry warping
    \item Motion bleeding across boundaries that breaks boundary coherence
\end{itemize}

\textbf{Rubric:}
\begin{itemize}[leftmargin=*,nosep]
    \item 5: Stable shape and boundaries across frames
    \item 4: Minor distortions without affecting the overall structure
    \item 3: Noticeable shape or boundary instability
    \item 2: Severe deformation or tearing across multiple frames
    \item 1: Structural collapse; the subject breaks apart or becomes unrecognizable
\end{itemize}
\end{minipage}
}
\captionof{figure}{\textit{Structural Consistency} evaluation criteria.}
\label{fig:appendix_structural_consistency}
\end{minipage}

\noindent\begin{minipage}{\linewidth}
\centering
\fcolorbox{black}{gray!10}{%
\begin{minipage}{0.95\linewidth}
\scriptsize
\textbf{Dynamic Consistency}

\textbf{Definition:} Temporal and physical coherence: whether motion and state transitions across frames are temporally stable and physically plausible.

\textbf{Anti-contamination rule:} Evaluate only time and physics coherence, including flicker, abrupt jumps, teleport-like changes, and physical implausibility. Ignore identity replacement and pure rendering artifacts unless they directly create apparent temporal flashing.

\textbf{Representative failure signs:}
\begin{itemize}[leftmargin=*,nosep]
    \item Flickering, flashing, or sudden global brightness or color shifts
    \item Abrupt appearance or position jumps
    \item Severe frame-to-frame jitter
    \item Gravity violations or impossible articulation
    \item Objects passing through each other
    \item Unrealistic acceleration or speed changes without a physical cause
\end{itemize}

\textbf{Rubric:}
\begin{itemize}[leftmargin=*,nosep]
    \item 5: Temporally stable and physically plausible dynamics
    \item 4: Minor temporal instability or minor physical oddities
    \item 3: Noticeable temporal instability or noticeable physical violations
    \item 2: Severe flicker, jumps, or severe physical violations
    \item 1: Dynamics critically broken
\end{itemize}
\end{minipage}
}
\captionof{figure}{\textit{Dynamic Consistency} evaluation criteria.}
\label{fig:appendix_dynamic_consistency}
\end{minipage}

\noindent\begin{minipage}{\linewidth}
\centering
\fcolorbox{black}{gray!10}{%
\begin{minipage}{0.95\linewidth}
\scriptsize
\textbf{Artifact Suppression}

\textbf{Definition:} Rendering-level defects independent of identity, geometry, or temporal coherence.

\textbf{Anti-contamination rule:} Evaluate only rendering defects such as blur, noise, grid artifacts, blockiness, and banding. Ignore identity replacement, shape deformation, and coherent motion unless the rendering artifact itself is the primary issue.

\textbf{Representative failure signs:}
\begin{itemize}[leftmargin=*,nosep]
    \item Blur, pixelation, or noise
    \item Texture melting or warping as a rendering defect
    \item Checkerboard, tiling, or grid artifacts
    \item Compression-like blockiness
    \item Color bleeding or unnatural banding
    \item Glitches unrelated to coherent motion, such as render seams or corruption
\end{itemize}

\textbf{Rubric:}
\begin{itemize}[leftmargin=*,nosep]
    \item 5: No visible artifacts
    \item 4: Minor artifacts
    \item 3: Noticeable artifacts affecting visual quality
    \item 2: Severe artifacts that strongly degrade viewing quality
    \item 1: Critically broken rendering
\end{itemize}
\end{minipage}
}
\captionof{figure}{\textit{Artifact Suppression} evaluation criteria.}
\label{fig:appendix_artifact_suppression}
\end{minipage}

\clearpage
\vfill

\end{document}